\pdfoutput=1

\documentclass[11pt]{article}

\usepackage[final]{acl}

\usepackage{times}
\usepackage{latexsym}
\usepackage{amsmath}
\usepackage{amssymb}
\usepackage{mathtools}
\usepackage{amsthm}
\usepackage{booktabs} 
\usepackage{graphicx, subfig}
\usepackage{hyperref}
\usepackage{url}
\usepackage{pgfplots}
\usetikzlibrary{pgfplots.groupplots}
\usepackage{wrapfig, lipsum, booktabs}
\usepackage{makecell}
\usepackage{xcolor}
\usepackage{amsfonts}
\usepackage{multirow}
\usepackage{tcolorbox}

\usepackage[T1]{fontenc}

\usepackage[utf8]{inputenc}

\usepackage{microtype}

\usepackage{inconsolata}

\usepackage{graphicx}
\usepackage{listings}
\usepackage{placeins} 
\lstdefinestyle{prompt}{
  basicstyle=\ttfamily\small,
  breaklines=true,
  frame=tb,
  breakindent=0pt,
  columns=fullflexible,
  backgroundcolor=\color[gray]{0.95},
  escapeinside={(*}{*)}
}

\lstdefinestyle{base}{
  language=C,
  emptylines=1,
  breaklines=true,
  basicstyle=\ttfamily\color{black},
  moredelim=**[is][\color{red}]{@}{@},
}
\def\BENCH{\textsc{GuideBench}}

\title{\textsc{GuideBench}: Benchmarking Domain-Oriented \\ Guideline Following for LLM Agents}

\author{
  Lingxiao Diao\textsuperscript{1,2,3}, Xinyue Xu\textsuperscript{1,3}, Wanxuan Sun\textsuperscript{3}, Cheng Yang\textsuperscript{3}\thanks{\ \ Corresponding authors. This work is partially supported by National Natural Science Foundation of China (62406188) and Natural Science Foundation of Shanghai (24ZR1440300).}, Zhuosheng Zhang\textsuperscript{1}\footnotemark[1] \\
  \textsuperscript{1}School of Computer Science, Shanghai Jiao Tong University \\
  \textsuperscript{2} Zhiyuan college, Shanghai Jiao Tong University\\
  \textsuperscript{3}ByteDance Inc \\
  \texttt{\{dlx421, xuxinyue2021, zhangzs\}@sjtu.edu.cn} \\
  \texttt{\{sunwanxuan, yangcheng.iron\}@bytedance.com}
}

\begin{document}
\maketitle

\begin{abstract}
Large language models (LLMs) have been widely deployed as autonomous agents capable of following user instructions and making decisions in real-world applications. 
Previous studies have made notable progress in benchmarking the instruction following capabilities of LLMs in general domains, with a primary focus on their inherent commonsense knowledge. 
Recently, LLMs have been increasingly deployed as domain-oriented agents, which rely on domain-oriented guidelines that may conflict with their commonsense knowledge. 
These guidelines exhibit two key characteristics: they consist of a wide range of domain-oriented rules and are subject to frequent updates.
Despite these challenges, the absence of comprehensive benchmarks for evaluating the domain-oriented guideline following capabilities of LLMs presents a significant obstacle to their effective assessment and further development. 
In this paper, we introduce \textsc{GuideBench}, a comprehensive benchmark designed to evaluate guideline following performance of LLMs. \textsc{GuideBench} evaluates LLMs on three critical aspects: (i) adherence to diverse rules, (ii) robustness to rule updates, and (iii) alignment with human preferences.
Experimental results on a range of LLMs indicate substantial opportunities for improving their ability to follow domain-oriented guidelines.
Data and code are available at \url{https://github.com/Dlxxx/GuideBench}.
\end{abstract}
\section{Introduction}
\label{sec:intro}

\begin{figure}[t]
    \centering
    \includegraphics[width=\columnwidth]{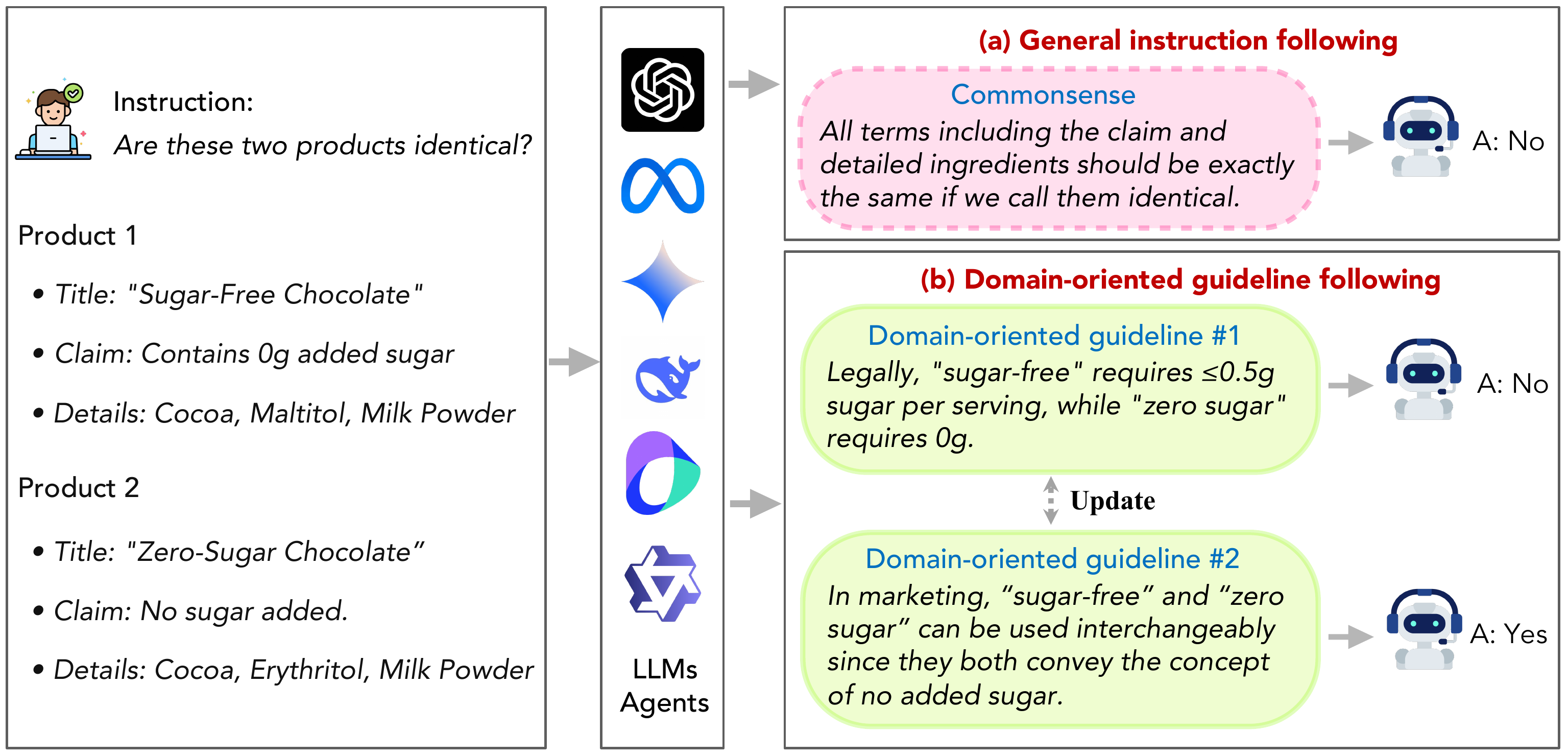}
    \caption{Comparison of general instruction following and domain-oriented guideline following. Given the instruction to determine whether two products are identical, previous benchmarks deems two products as not identical based on LLMs' commonsense. However, distinct domain knowledge reflected by the update of the guidelines (\#1$ \rightarrow$ \#2) may yield different conclusions. 
    }
    \label{fig:intro}
    \vspace{-5mm}
    \end{figure}

The advancement of large language models (LLMs) has driven the development of autonomous agents capable of performing complex real-world tasks without manual intervention, including operations and maintenance ~\cite{chen2024revolutionizing} and incident management ~\cite{roy2024exploring}. A key factor in this progress lies in the instruction following capability, which determines how effectively LLMs align with human intents ~\cite{ouyang2022training}.

Compelling evidence suggests that current LLMs still struggle with instruction following ~\cite{yin2023llm,mu2023can,heollms}. To evaluate this capability, significant advancements have been made in previous studies concerning both generic instructions ~\cite{lialpacaeval,zheng2023judging} and compositional instructions ~\cite{wen2024complexbench,sun2024beyond,qin-etal-2024-infobench,jiang-etal-2024-followbench,he2024can}. For example, ComplexBench ~\cite{wen2024complexbench} employs complex instructions with multiple constraints, while RuleBench ~\cite{sun2024beyond} incorporates inferential rules. However, these studies primarily target general domains and reflect commonsense knowledge within LLMs.

Beyond general instruction following, a significant challenge lies in the ability of LLMs to follow domain-oriented guidelines, especially as these models are increasingly deployed as domain-oriented agents (Figure \ref{fig:intro}).\footnote{\textit{Instructions} refer to specific, direct tasks that agents are expected to perform. In contrast, \textit{guidelines} consist of domain-specific rules, which may occasionally be counterfactual, and are tailored to address user needs. Guidelines often serve as supplementary resources to assist in the completion of instructions within particular domains.} 
Guideline-following exhibits two key characteristics:
(i) \textbf{domain rules}: guidelines consist of numerous rules grounded in domain knowledge, which may involve compositional, conditional, or nested relationships;
(ii) \textbf{frequent updates}: guidelines are subject to frequent updates to stay aligned with evolving standards and regulations.
Additionally, the guidelines may occasionally conflict with the commonsense knowledge possessed by the LLM~~\cite{xu-etal-2024-knowledge-conflicts,xieadaptive}.
Therefore, ensuring adherence to domain-oriented guidelines poses a significant challenge.

This paper introduces \BENCH{}, a comprehensive benchmark to evaluate guideline following performance of LLMs in real-world applications. \BENCH{} is developed through a combination of automatic synthesis and human-in-the-loop refinement, resulting in 1272 instances across 7 distinct categories, namely audit algorithm, price matching, text relevance, math, agent chatting, summarization and hallucination detection. The benchmark evaluates LLMs across three key dimensions: (i) adherence to domain rules, (ii) robustness to rule modifications, and (iii) alignment with human preferences.

Based on \BENCH{}, we conduct a comprehensive evaluation of 18 prominent LLMs commonly used as the backbone of agent systems.
Concretely, we provide both multiple options and question-answering problems to better align with real-world applications. The evaluated LLM is tasked with selecting the optimal response or generating the correct answer based on external rules, rather than relying on its commonsense knowledge.
Our results highlight considerable potential for improving the domain-oriented instruction following capabilities of LLMs. 
Notably, most LLMs fall below 60\% on guideline following in math tasks---except for Deepseek-R1, which achieves 65.38\%.
Analysis highlights the critical role of guidelines and substantial gain brought by chain-of-thought (CoT) in complex tasks. Error analysis demonstrates considerable room for improvement in domain-specific guideline following of current LLMs. Our contributions are as follows:

\begin{enumerate}
    \item We introduce \BENCH{}, a comprehensive benchmark to evaluate guideline following capability of LLMs in real-world applications. Compared to the previous benchmarks, the guidelines in \BENCH{} are tailored for specific domains rather than generic rules for all scenarios, providing a more accurate measure of LLMs' capability to follow user needs with external domain knowledge.
    \item We propose a pipeline that automates both synthesis of domain-oriented guideline rules and evaluation of curated multiple options tasks and question-answering tasks.  
    \item We evaluate LLMs using \BENCH{}, showing that they struggle to effectively follow complex domain-oriented rules. In addition, we examine the importance of domain guidelines and CoT and conduct in-depth error analysis, offering valuable insights to facilitate future research.
\end{enumerate}

\section{Related work}
Our work falls into the field of LLM-based autonomous agents. We first review recent advances in developing such agents, followed by an exploration of instruction following evaluation.

\subsection{LLM-based Autonomous Agents}
Driven by innovations in prompt engineering, imitation learning and reinforcement learning, LLM-based autonomous agents have evolved rapidly, expanding their capabilities to perform complex tasks across various domains. Building intelligent agents that can autonomously learn and act in dynamic environments has long been a critical goal of AI research \citep{searle1969speech, maes1995agents, hendler1999there, wang2023survey, xi2023rise, zhou2023agents}.

LLM-based agents have found applications in fields ranging from engineering \citep{li2023camel, mehta2023improving, qian2023communicative} and natural sciences \citep{bran2023chemcrow, kang2023chatmof, boiko2023emergent} to social sciences \citep{aher2023using, akata2023playing, ma2023understanding, dan2023educhat}, where they execute tasks based on language instructions in real-world or simulated environments.

Among these agents, operational agents~\citep{chen2024revolutionizing,roy2024exploring} stand out for their potential to enhance the efficiency of critical tasks, such as system analysis, auditing, and process optimization. Their effectiveness hinges on their ability to accurately follow user intent, ensuring that actions are aligned with domain-specific instructions. 
This capability is essential for improving both the effectiveness and reliability of operational workflows, which is the central focus of this paper.

\subsection{Instruction Following Benchmarks}
\label{subsec:benchmark}
Instruction following plays an important role in determining the practicality of modern LLMs. Numerous attempts have been made to evaluate it from various aspects.
Earlier work focuses on simple human instructions formed with a single constraint, such as semantic~\citep{zheng2023judgingllmasajudgemtbenchchatbot, dubois2024lengthcontrolledalpacaevalsimpleway, liu2024alignbenchbenchmarkingchinesealignment} and format constraints~\citep{zhou2023instructionfollowingevaluationlargelanguage, xia2024fofobenchmarkevaluatellms, tang2024strucbenchlargelanguagemodels}.

As LLMs gradually serve to address real-world tasks, regular or industrial users form more complex instructions, which naturally calls for the evaluation of complex instruction following~\cite{jiang-etal-2024-followbench,qin2024infobenchevaluatinginstructionfollowing}.
Notbaly, WizardLM~\cite{xu2023wizardlmempoweringlargelanguage} employs two strategies, \textit{in-breadth evolving} and \textit{in-depth evolving}, to form complex instructions from simple ones.
CELLO~\cite{he2024largelanguagemodelsunderstand} defines complex instructions from task descriptions and input text, and evaluates LLMs with real-world scenarios data.
COMPLEXBENCH~\cite{wen2024complexbench} manually synthesizes complex instructions from reference instructions collected in real-world applications and existing benchmarks. 
However, current benchmarks neglect to model the complexity and adaptability of domain-specific tasks, which are critical factors in complex instructions and pose structural challenges to evaluating LLMs.

\section{The \BENCH{} Benchmark}
\label{sec:benchmark_creation}

\begin{figure*}[t]
    \centering
    \includegraphics[width=\linewidth]{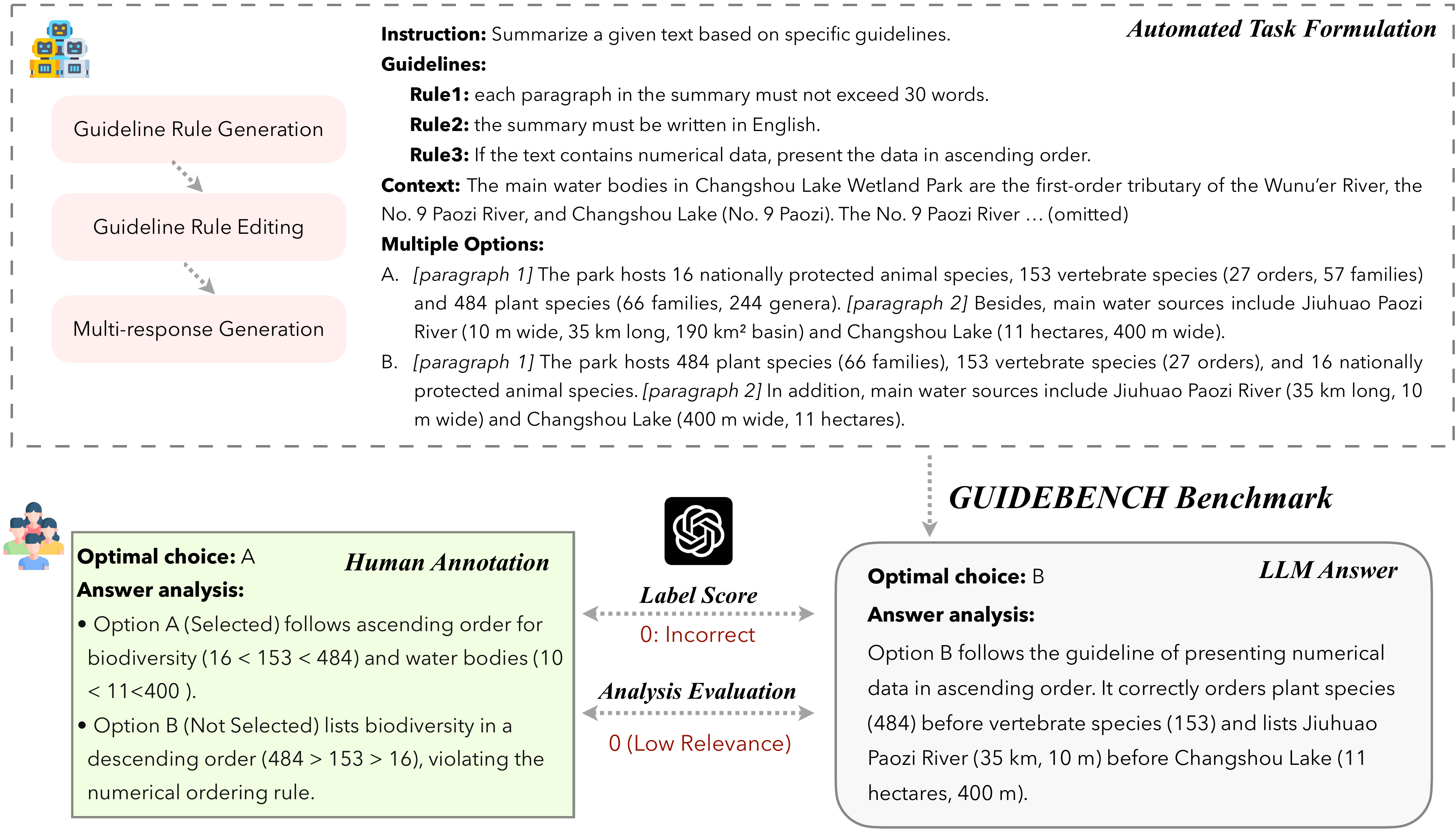}
    \caption{Illustration of a sample task in \BENCH{}. A typical multiple options task includes four main components: 1) Instruction, the overarching task objective; 2) Guidelines, a set of domain-specific rules that inform the task structure; 3) Context, a relevant text passage; and 4) Multiple Options, omitted for question-answering tasks, a set of diverse responses generated by LLMs. After LLMs generating the answer and corresponding analysis, \BENCH{} evaluates the correctness of the results based on human-annotated data, with answer analysis providing some level of interpretability.}
    \label{fig:framework}
    \vspace{-5mm}
\end{figure*}

\begin{figure}[t]
    \centering
    \includegraphics[width=\columnwidth]{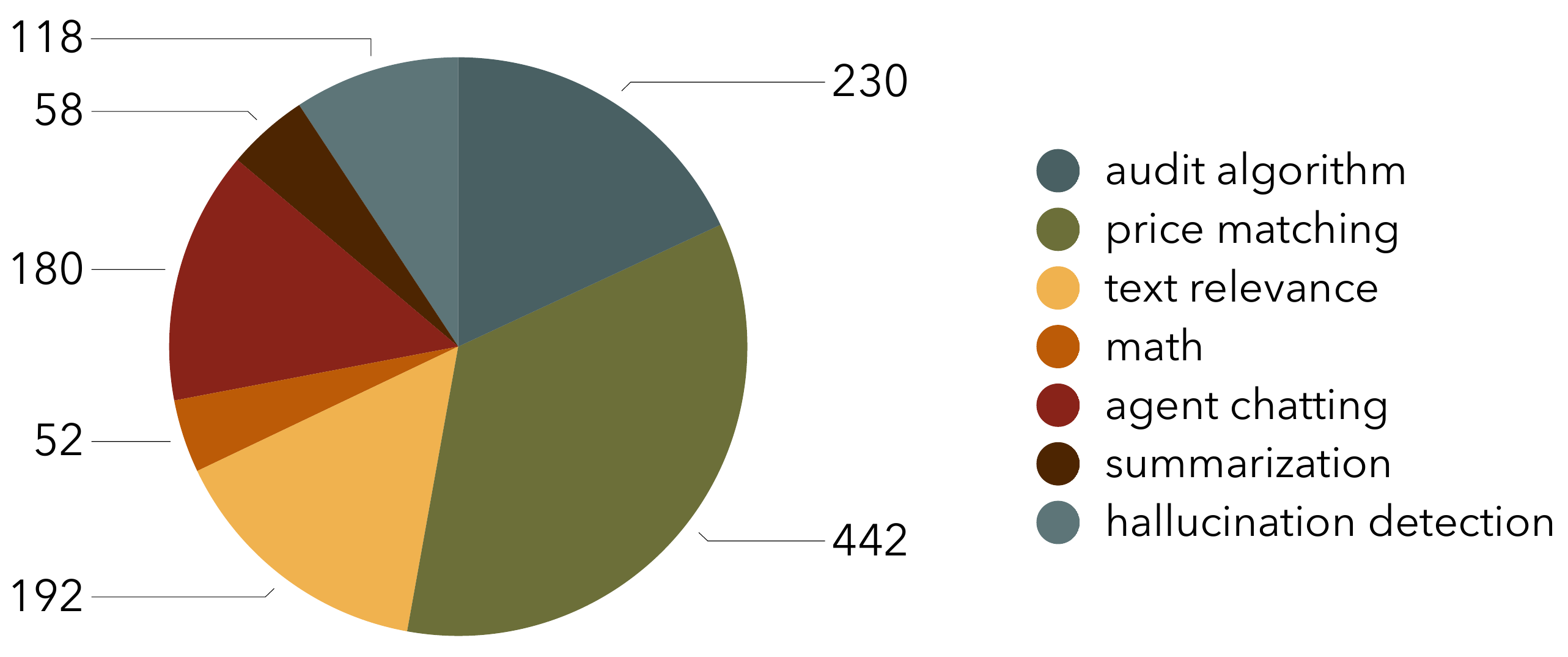}
    \caption{Dataset distribution of \BENCH{}, which consists of 1272 tasks across 7 categories. The proportion of each domain is shown in the graph.}
    \label{fig:benchmark}
    \vspace{-5mm}
\end{figure}

\subsection{Overview of \BENCH{}}
\label{sec:overview_sub}
We introduce \BENCH{}, a comprehensive benchmark designed to evaluate the guideline following capability of LLMs. \BENCH{} spans 7 typical categories: \textit{audit algorithm}, \textit{price matching}, \textit{text relevance}, \textit{math}, \textit{agent chatting}, \textit{summarization} and \textit{hallucination detection} with the total number of 1272 tasks. The detailed statistics are detailed in Figure \ref{fig:benchmark}. Each category includes rigorously modified tasks, created through rule-based transformations to simulate changes in real-world guidelines. 

Specifically, \BENCH{} is designed to evaluate distinct aspects of LLM performance: \textit{audit algorithm} tasks examines models' procedural compliance; \textit{price matching} tasks test models’ ability to adapt to dynamic e-commerce market data; \textit{text relevance} tasks assess semantic alignment precision; \textit{math} tasks challenge logical reasoning; \textit{agent chatting} tasks assess conversational coherence across multiple turns; \textit{summarization} tasks evaluate the quality of information distillation under constrained conditions; and \textit{hallucination detection} measures factual consistency. Together, these categories form a comprehensive framework for profiling LLM capabilities.

The data construction process follows three phases: initial data collection from reliable domain sources from real-wold applications, followed by guideline generation using scenario simulation templates and multi-response generation, and final quality verification through baseline model validation and expert review. Detailed prompt templates are documented in Appendix \ref{app:all_prompts} for reproducibility.

\BENCH{} serves as a diagnostic tool to identify LLM vulnerabilities when facing evolving operational standards, offering valuable insights for improving the robustness of LLMs. Its modular structure enables both comprehensive evaluation and targeted analysis of domain-specific performance, making it especially useful for detecting weaknesses in commercial LLM deployments.

\subsection{Data Curation Process}
As shown in Figure \ref{fig:framework}, each task in \BENCH{} comprises several key components: 

\begin{itemize}
    \item \textbf{Instruction}: the overarching task objective.
    \item \textbf{Guidelines}: a set of user-defined rules incorporating domain knowledge, including multiple guideline rules.
    \item \textbf{Context}: a relevant text passage, such as product information in price matching, problem descriptions in math and user-agent dialogues in agent chatting.
    \item \textbf{Multiple Options}: a set of quality-diverse responses generated by LLMs (optional).\footnote{For agent chatting, summarization and hallucination detection tasks, we provide multiple options; while for audit algorithm, price matching, text relevance and math tasks, we structure them as question-answering problems to better align with real-world applications. Task details will be provided in Section \ref{sec:formats}.}
\end{itemize}

The following subsections outline the four stages involved in the automatic construction of the \BENCH{} dataset: (i) Data Collection, (ii) Guideline Rule Generation, (iii) Guideline Construction, and (iv) Multi-response Generation. Then it is followed by a discussion on Data Quality Control, which encompasses two critical processes: LLM filtering and Human Annotation.

\subsubsection{Data Collection} 
Our data collection process follows three key stages. First, we identify the categories that are most beneficial for operational applications, with a focus on those that remain under-explored in current research. Based on this principle, we prioritize 7 typical categories: audit algorithm, price matching, text relevance, math, agent chatting, summarization and hallucination detection. Then, we manually extract several seed instructions from practical use cases within these categories, and subsequently derive domain-specific instructions and basic guidelines, prompts seen in Appendix \ref{app:grg}. 
These seed instructions are then expanded through rule generation process as detailed in Section \ref{para:rule_generation}, utilizing LLMs to create a large set of domain-specific guidelines. 
After generating the guidelines, we use LLMs to produce the context for each task uniformly, prompts seen in Appendix \ref{app:gc}.

\subsubsection{Guideline Rule Generation}
\label{para:rule_generation}
We begin by extracting key elements based on a system prompt that includes the overall task objectives, input and output specifications, and detailed requirements for rule construction. Specifically, the input specifications define the types of data accepted by the system, while the output specifications describe the expected format and content of the results, such as classifications, summaries, or actionable recommendations. Additionally, the system prompts also provide detailed requirements for rule construction, such as requirements for rule content, covered scenarios, and diversity. The system prompt is provided in Appendix \ref{app:grg}.

Next, we extract key elements from the task guideline, including the task objectives, input and output specifications, and rule set requirements. Based on the extracted elements, we proceed to draft the initial rule set. Each rule is structured with a condition part that defines the triggering conditions and an operation part that specifies the actions to be taken. We then categorize the rules, introducing new categories to enhance the diversity and applicability of the rule set, with detailed prompts in Appendix \ref{app:ptgr}. 

Once the rule set is generated, we perform an automatic quality inspection process. Specifically, we leverage GPT-4o~\cite{openai2024gpt4technicalreport} to eliminate duplicate rules, as detailed in Appendix \ref{duplicate}, followed by a manual review to ensure logical consistency and identify any unrealistic or implausible cases. This process results in a diverse collection of 537 guideline rules.

\subsubsection{Guideline Construction}
\label{subsub:guidelineconstruction}
After generating individual guideline rules, we extract and assemble them into comprehensive guidelines using three distinct methods to enhance diversity:

\begin{itemize}
    \item \textbf{Random Selection}: Randomly selecting $k$ guideline rules within the same domain to form a guideline.
    \item \textbf{Diversity-based Selection}: Prioritizing the selection of guideline rules from different types within the same domain to ensure diversity in the assembled guideline.
    \item  \textbf{Semantic-based Selection}: Leveraging an LLM to choose appropriate guideline rules based on the overarching instruction, ensuring semantic coherence in the assembled guideline. The prompt is given in Appendix \ref{app:grr}.
\end{itemize}

Besides, for domain-oriented agents, user requirements are constantly evolving, leading to continuous updates in guidelines that LLM agents follow. Such modifications can fundamentally alter the generated responses. Enhancing the robustness of LLM agents in following updated guidelines is thus crucial. To achieve this, we employ LLMs to modify the rules within the guidelines, ensuring that the generated responses align with the intended changes, prompts seen in Appendix \ref{app:edit_rules}.

\subsubsection{Multi-Response Generation}

Lastly we construct high-quality multiple-choice questions based on the generated guideline described in Section \ref{subsub:guidelineconstruction}. Specifically, we incorporate generated guidelines as part of the prompt, seen in Appendix \ref{app:gc}, and employ LLMs
to generate relevant text contexts. Then, we assemble questions by integrating the generated contexts with the corresponding guidelines. Finally, we prompt LLMs to generate answer options, resulting in a well-structured multiple-choice format in Appendix \ref{app:gq}.

\subsubsection{Data Quality Control}

For domain-specific models, generating content that aligns with user preferences is essential \cite{sun2024beyond}. To ensure the quality and accuracy, we employed a combination of automated and manual methods. In the Guideline Rule Generation phase in Section \ref{para:rule_generation}, LLM filtering was initially applied to eliminate duplicate or low-quality rules. Subsequently, after task generation, LLMs were utilized to generate both the Optimal Option and the Answer Analysis. Detailed prompts for these steps are provided in Appendix \ref{app:gq}.

Alongside the automated LLM-based approach, human annotations are performed by experts with specialized knowledge in the categories outlined in Section \ref{sec:overview_sub}, and academic backgrounds in AI and computer science. 
The annotation process involved a comprehensive review of tasks based on the LLM-generated labels. Optimal Option labels for multiple options tasks and Reference Answer labels for question-answering tasks were carefully corrected in accordance with the established guidelines, and ambiguities in the corresponding Answer Analysis or Reference Analysis were resolved.

\section{Task Formulation}
In this section, we present the task format of \BENCH{} and outline the evaluation criteria.

\begin{table*}[!htb]
  \centering
  \renewcommand\tabcolsep{3.6pt} 
  \small
  {
  \begin{tabular}{l|c|ccccccc} 
  \toprule
  \multirow{3}{*}{Models} & \multirow{1}{*}{\textit{All}} & \multicolumn{7}{c}{\textit{Task Categories}} \\
    \cmidrule{2-2}
    \cmidrule(lr){3-9}
    & \multirow{2}{*}{Accuracy} & Audit & Price &  Text  & \multirow{2}{*}{Math} & Agent & \multirow{2}{*}{Summarization} &  Hallucination \\
    &  & Algorithm & Matching &  Relevance &  & Chatting &  &  Detection \\
    
    \midrule
    
    o1  & 79.17  & 73.48  & 76.24 & 79.69  & \underline{48.08}  & 92.78 & 81.03 & 92.37 \\
    GPT-4o & \underline{86.48} & \underline{96.52}  & 84.84  & 81.25  & 13.46  & \textbf{100} & 82.76 & 94.92  \\
    
    GPT-4o* & 80.90 & 94.78  & 74.66  & 80.21  & 7.69  & 95.56 & 68.97 & 94.07  \\\midrule

    Deepseek-R1  & \textbf{87.26}  & 93.04  & 80.32 & 84.90 & \textbf{65.38} & \underline{98.89}  & \textbf{89.66 } & \textbf{96.61} \\
    Deepseek-V3 & 83.96 & \textbf{97.39}  & \textbf{91.18}  & 53.65  & 5.77 & \underline{98.89} & 77.59 & 94.92  \\ \midrule

    Gemini2.5-pro-exp & 80.9 & 90 & 75.79 & \underline{85.94} & 44.23 & 80.00 & 87.93 & 93.22 \\  \midrule
    
    Mistral-7B-Instruct & 69.58 & 86.52 & 66.06 & 77.60 & 1.92 & 58.33 & 58.62 & 88.98 \\ \midrule
    
    Yi-1.5-6B & 56.05 & 50.43 & 66.29 & 43.75 & 7.69 & 66.11 & 20.69 & 72.03 \\        
    Yi-1.5-34B & 72.64 & \textbf{97.39} & 77.60 & 67.71 & 25.00 & 60.00 & 56.90 & 61.86 \\  \midrule

    Gemma-3-4b-it & 61.71 & 58.70 & 56.11 & 75.00 & 0 & 76.67 & 72.41 & 66.10 \\ \midrule
    
    QwQ-32B-Preview & 71.15 & 53.04 & 64.71 & 82.29 & 28.85 & 94.44 & 77.59 & 92.37\\ 
    R1-Distill-Qwen-7B  & 57.94 & 90.43 & 65.16  & 4.69  & 17.31  & 51.11  & 58.62  & 82.20 \\
    R1-Distill-Qwen-32B  & 85.69 & 93.48 & \underline{86.65}  & 78.12  & 7.69  & 98.33  & \underline{84.48}  & 94.92 \\
    
    Qwen2.5-7B  & 42.61 & 75.65  & 54.30  & 66.67 & 0.00 & 0.00  & 0.00  & 0.00 \\
    Qwen2.5-7B-Instruct  & 81.13  & \textbf{97.39} & 74.43 & 81.77  & 1.92  & 96.11  & 67.24  & 92.37 \\
    Qwen2.5-32B  & 83.73  & 91.30 & 81.67 & 80.73 & 3.85  & \textbf{ 100}  & 77.59  & 94.92 \\\midrule
    Vicuna-7B  & 33.57  & 60.87  & 48.87 & 30.73 & 0.00 & 0.00 & 20.69  & 0.00 \\\midrule

    Llama3-8B-Instruct  & 27.36  & 61.74  & 30.77  & 34.90  & 0.00 & 0.56 & 3.45 & 0.00 \\
    Llama-3.3-70B-Instruct & 86.24 & \textbf{97.39} & 82.58 & \textbf{86.98} & 11.54 & 96.67 & 82.76 & \underline{95.76} \\ 
    \bottomrule
  \end{tabular}}
  \caption{Main results (\%) of different models across 7 task categories. All answers are parsed by GPT-4. Accuracy is computed based on the number of correctly predicted labels. Segment 1: GPT series; Segment 2: Deepseek series; Segment 3: Gemini series; Segment 4: Mistral series; Segment 5: Yi series; Segment 6: QwQ series; Segment 7: Gemma series; Segment 8: Qwen series; Segment 9: Vicuna series; Segment 10: Llama series. The best-performing results are shown in \textbf{bold} face, and the second-best are \underline{underlined}. GPT-4o* denotes the ablation study using GPT-4o with basic \textit{Instruction} and \textit{Context}, but without \textit{Guidelines}.}
  \label{tab:results}
  \vspace{-5mm}
\end{table*}

\subsection{Task Format}\label{sec:formats}
\BENCH{} includes a comprehensive evaluation framework based on two core components: task composition and evaluation protocol.
Note that open-ended tasks like agent-chat or text summary remain challenging to directly evaluate, we adopt question-answering (QA) and multiple options structures to enhance the difficulty and diversity of the dataset.

As depicted in Figure~\ref{fig:framework}, a task is represented as either multiple options structure~$(\mathcal{I}, \mathcal{G}, \mathcal{C}, \mathcal{M})$ or QA structure~$(\mathcal{I}, \mathcal{G}, \mathcal{C})$, where $\mathcal{I}$ refers to Instruction, $\mathcal{G}$ to Guidelines including the set of guiding rules, $\mathcal{C}$ to Context, and $\mathcal{M}$ to multiple answer options generated by LLMs. Thus, we define the formulated task as:
$
    T = (I, G, C, M)
$
or $
    T = (I, G, C).
$

When performing the task, the LLM is required to first analyze the Context based on the Instruction and Guidelines, resulting in an \textit{analysis}. Subsequently, the model generate answers based on the \textit{analysis}, producing the final \textit{answer}.
The task formulation is:
$
    f: p_{\theta} (T) \to (\textrm{analysis}, \textrm{answer}),
$
where $p_{\theta}(\cdot)$ denotes the language model mapping the task $T$ to $\textit{analysis}$ and an $\textit{answer}$ response.

\subsection{Evaluation Criteria}
\label{subsec:criteria}
Based on the dataset, we evaluate the accuracy of public LLMs in following given instructions. Adhering to external domain knowledge demands both rule comprehension and rule application capabilities of LLMs. 

To assess their proficiency in guideline adherence, we design two evaluation recipes.
As illustrated in Figure \ref{fig:framework}, the Evaluation stage consists of label scoring, verifying the correctness of the generated answer, with human annotation as ground truth.
First, LLMs are presented with task specifications and prompted to generate analysis, formalized as $p_{\theta}(T) \to \textrm{analysis} \label{eq1}$. Then, based on task specifications and analysis from the first step, LLMs are required to generate the final answer, formalized as $p_{\theta} (T, \textrm{analysis}) \to \textrm{answer}\label{eq2}$.

\paragraph{Label Score for Candidate Answer.} Since both QA and multiple options questions exist, to evaluate the ability of LLMs to accurately adheres to the specified guidelines, a label-based test compares the model-generated answers with the reference answer or optimal choice from the consensus of human annotators.

\section{Experiments}
\label{sec:experiment}
In this section, we describe experimental setup, evaluate 18 popular LLMs and present main results.

\subsection{Setup}
\label{subsec:setup}
\paragraph{Baselines.} We comprehensively assess 18 LLMs, including API-based models and open-source models. The API-based models include GPT series~\cite{openai2024gpt4technicalreport} and Deepseek series~\citep{deepseekai2024deepseekv3technicalreport}.
The open-source models include Llama series~\cite{grattafiori2024llama3herdmodels}, Qwen series~\cite{qwen2.5} and Vicuna series~\cite{vicuna2023}.
See Table \ref{tab:LLMs_evaluated} and complete experimental settings including hyper-parameters and costs are in Appendix \ref{app:Experiment_Settings}.

\paragraph{Prompt Setting}
Following \citet{yuan2024rjudgebenchmarkingsafetyrisk}, we adopt Zero-Shot-CoT prompting \cite{kojima2023largelanguagemodelszeroshot} to induce LLMs to generate the reasoning steps before producing the final answer. This kind of analyze-then-output process has been shown to improve reasoning performance, as well as interpretability \cite{zhang2023ignitinglanguageintelligencehitchhikers}.
See Appendix \ref{app:aq} for details.

\paragraph{Metrics}
As shown in Figure~\ref{fig:framework}, given human-annotated data as the ground truth, the evaluation process involves determining whether the LLM-generated Candidate Answer is correct. Given that \BENCH{} covers seven categories, we measure overall correctness using accuracy, while also calculating accuracy, Recall, and Precision score for each individual domain. It allows to assess both the overall performance and domain-specific results.

\paragraph{Response Parsing}
During experiments, we observed that many baseline models, particularly those with smaller parameter sizes, failed to strictly adhere to the required output format. The lack of consistency significantly affected the experimental results. To mitigate this issue, we leveraged GPT-4 to parse and reformat model responses into the expected format, with corresponding prompts provided in Appendix \ref{Extracting}. Pre-processing data enabled a more reliable evaluation by standardizing the responses, empowering to compute scores effectively. 

\subsection{Main Result}
\label{subsec:main_result}

We present parsed and raw results in Table ~\ref{tab:results} and Table~\ref{tab:not_parsed_main_results}, respectively. Based on these results, we have the following key findings.

(i) Most of the LLMs struggle significantly on \BENCH{}, with their accuracy falling below 30\% when formatting inconsistencies are accounted for. Even after normalizing format variations, model performance exhibits substantial disparities, highlighting the uneven capabilities across different models. While GPT-4o demonstrates strong overall proficiency with an accuracy of 86.48\%, its severe weakness in math (13.46\%) remains a critical limitation. In contrast, the Deepseek series showcases domain-specific strengths: Deepseek-R1 achieves the highest overall accuracy (87.11\%) and leads in math with 61.54\%, while Deepseek-V3 outperforms in tasks requiring precision, such as price matching (91.18\%) and audit algorithm verification (97.39\%). The Qwen models reveal a parameter-dependent trend, with the 32B variant performing on par with GPT-4o in several areas, whereas the 7B versions lag considerably. Meanwhile, Vicuna-7B and Llama3-8B-Instruct exhibit pronounced deficiencies, particularly in math and agent chatting tasks, underscoring their limitations relative to other model families.

(ii) Task-specific results reveal even more pronounced performance variations across different categories. While some models achieve near-perfect accuracy in tasks such as audit algorithm and agent chatting—demonstrating strong contextual comprehension—math emerges as the most challenging domain. Most LLMs fail to surpass 60\% in guideline following for math tasks, with Deepseek-R1 being the only exception, reaching 65.38\%. Tasks like price matching and text relevance reveal highly competitive performance, with Deepseek-V3 setting the benchmark for both. Similarly, hallucination detection and summarization tasks highlight notable discrepancies, with Deepseek-R1 consistently outperforming its peers. These findings illustrate the distinctive strengths and weaknesses of each model family while reinforcing the persistent difficulties in mathematical reasoning.
\section{Analysis}
\label{sec:analysis}
In this section, we conduct a comprehensive analysis of experimental results to investigate the impact of various mechanisms on model performance. 

\paragraph{Importance of Domain-Specific Guidelines}

We conduct ablation study experiment using GPT-4o with basic Instruction and Context but without \textit{Guidelines}, dubbed as GPT-4o* in Table~\ref{tab:results}, using prompts provided in Appendix \ref{app:without_guide}.
The omission of Guidelines leads to a noticeable accuracy decline in price matching, math, agent chatting and summarization tasks, suggesting that guideline rules plays a pivotal role in aligning model outputs more closely with dynamic and diverse domain requirements. 
Thus, \textit{Guidelines} contribute not only to improving accuracy but also to maintaining task-specific coherence and reliability, underscoring their importance for achieving robust model performance across various tasks.

\begin{figure}[t]
    \centering
    \subfloat[Model performance on precision score.]{%
        \includegraphics[width=\columnwidth]{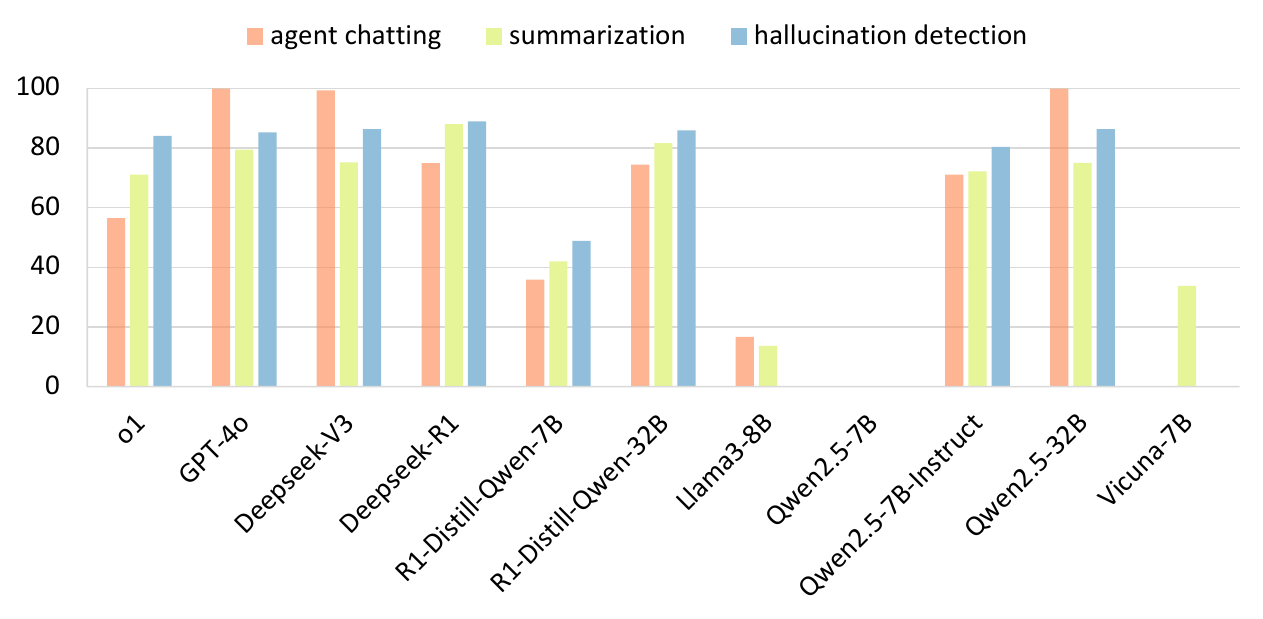}
        \label{fig:result_precision}
    }
    \vspace{0mm}
    \subfloat[Model performance on recall score.]{%
        \includegraphics[width=\columnwidth]{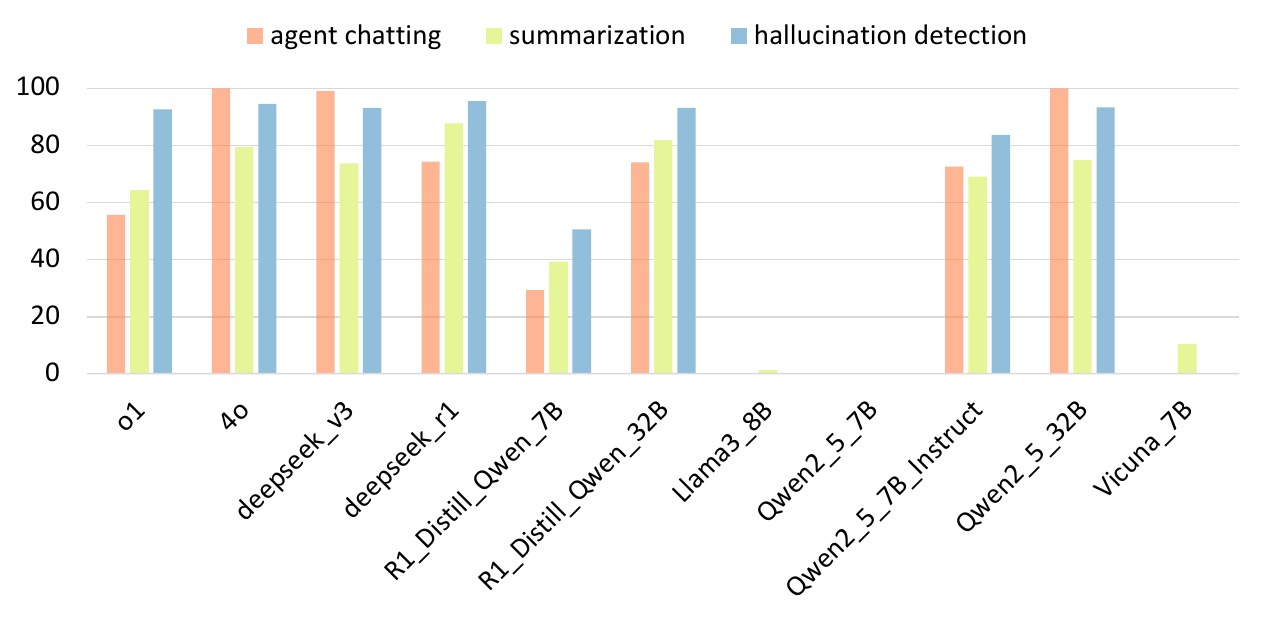}
        \label{fig:result_recall}
    }
    \vspace{-2mm}
    \caption{Model performance in agent chatting, summarization and hallucination detection tasks.}
    \vspace{-5mm}
\end{figure}

\paragraph{Performance and Accessibility Trade-offs}

In multiple-option scenarios of agent chatting, summarization, and hallucination detection tasks, the model performance in terms of precision and recall is shown in Figures~\ref{fig:result_precision} and~\ref{fig:result_recall}, respectively. 
We see that the GPT and Deepseek series models lead by a significant margin. 
However, challenges arise in operational scenarios, where closed-source models excel but are not available for domain-specific applications. On the other hand, while open-source models, such as Qwen series, are accessible, their performance still exhibits substantial variability, highlighting a notable gap in quality and stability.

\paragraph{Influence of CoT}
\label{para:influ-cot}
To analyze the impact of CoT outputs, we also prompt Deepseek-R1, the best-performing model, to output pure results without CoT in summarization and math tasks, as detailed in Appendix~\ref{app:re_evaluation_prompt}.

\begin{table}[htb]
\centering
    \renewcommand\tabcolsep{4pt} 
    \scalebox{0.7}{\begin{tabular}{l|ccc|ccc}\toprule
    & \multicolumn{3}{c|}{math}   & \multicolumn{3}{c}{summarization} \\ \cmidrule{2-7} 
            & Accuracy   & Precision  & Recall     & Accuracy  & Precision & Recall    \\ \midrule
    w/ CoT  & \textbf{65.38} & \textbf{47.92} & \textbf{49.22} & \textbf{89.66} & 88.04 & 87.73  \\ \midrule
    w/o CoT &  42.31 &  26.22  & 26.00 & \textbf{89.66} & \textbf{88.59} & \textbf{90.45} \\ \hline
    \end{tabular}}
    \caption{Result(\%) of metrics in math and summarization tasks. Better results are in \textbf{bold} face.}
    \label{tab:both}
    \vspace{-5mm}
\end{table}

As shown in Table~\ref{tab:both}, we see that CoT provides substantial improvements for complex tasks, but offers little benefit for simpler ones.
Specifically, the difference of model performance for summarization tasks is below 3\%, whereas for math tasks, it exceeds over 20\%. 

Meanwhile, we have identified CoT reasoning patterns in the model that align with phenomena discussed in recent literature: over-thinking~\cite{chen2025think23overthinkingo1like} and under-thinking~\cite{wang2025thoughtsplaceunderthinkingo1like}.

Over-thinking occurs when the model generates redundant problem-solving approaches for simple tasks.
For instance, in a price matching task in Figure~\ref{fig:over-think} in Appendix~\ref{app:overandunder}, where the goal is to determine whether Product A and Product B are the same, we observed that Deepseek-R1 often exhibited excessive reasoning through multiple self-corrections (e.g., \textit{"Wait, but..."}, \textit{"However, wait..."}). 
While ultimately producing correct answers, this verbosity resulted in 62-second latency and token redundancy.
Under-thinking emerges conversely in complex math reasoning, where frequent shifts in reasoning prevent the model from adhering to a consistent and correct exploration path.
For instance, in a math task in Figure~\ref{fig:underthinking} in Appendix~\ref{app:overandunder}, where the goal is to solve coupon math calculation problems, the model failed to figure out the best solution, neglecting the available 22 CNY coupon.

These dual phenomena manifest distinct characteristics across different task complexities, introducing optimization challenges in reasoning efficiency and depth control. Our findings underscore the need for adaptive reasoning mechanisms that dynamically adjust cognitive effort based on task complexity and reward signals.

\begin{table}[t]
\centering
\scalebox{0.7}{\begin{tabular}{cccccc}
\toprule
Original & PoT & Equivalent Math & Rules Reorder & 2-shot Demo \\
\midrule
65.38 & 30.27 & \textbf{86.54} & 65.38 & 65.38 \\
\bottomrule
\end{tabular}}
\caption{
Accuracy Results (\%) of different math task formulation strategies tested on Deepseek-R1. The best result is in \textbf{bold} face.
}
\vspace{-5mm}
\label{tab:w1}
\end{table}

\begin{figure}[t]
    \centering
    \includegraphics[width=\columnwidth]{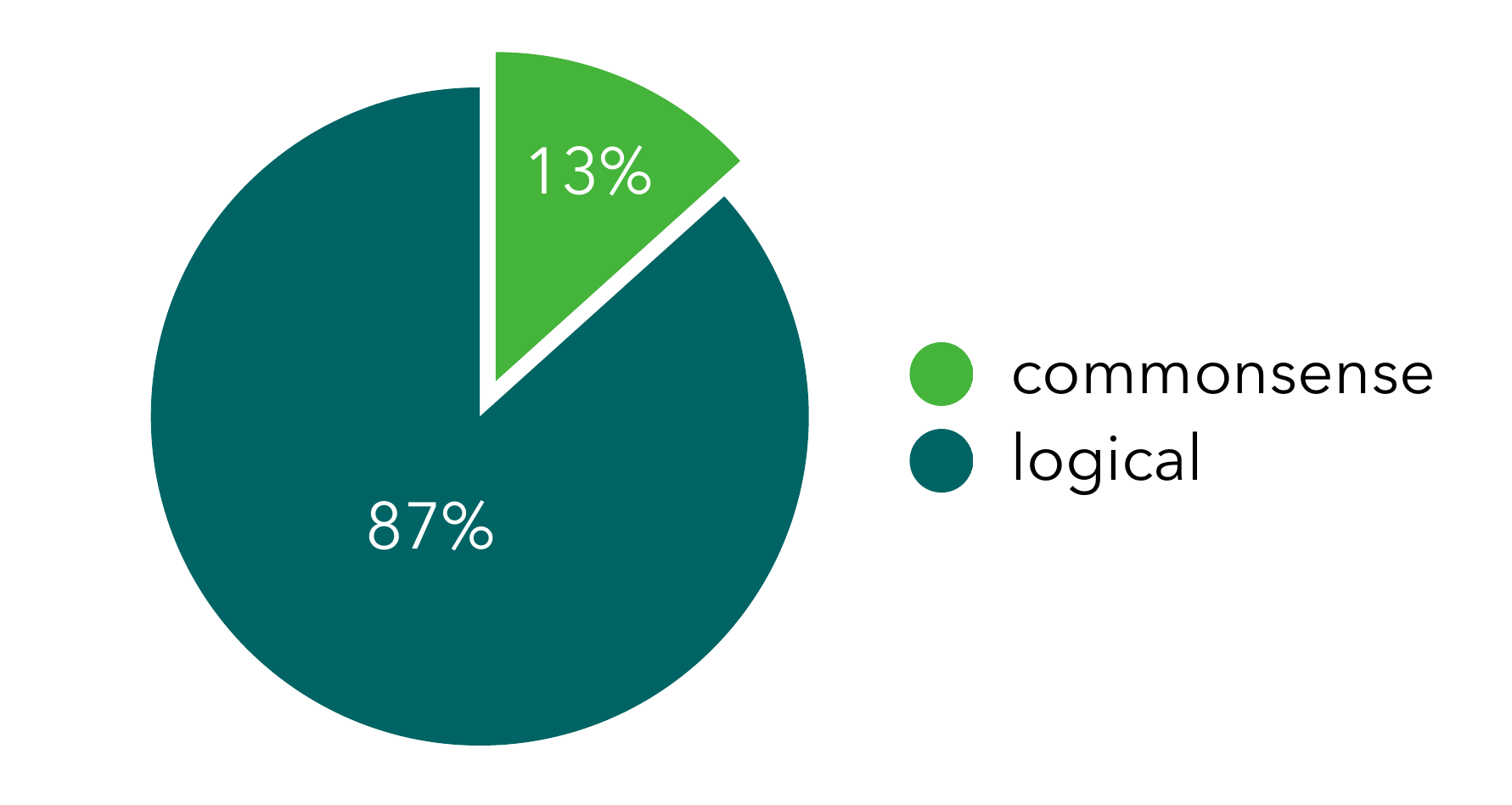}
    \caption{Categorization analysis}
    \label{fig:error_portion}
    \vspace{-5mm}
\end{figure}

\paragraph{Error Analysis in Math Tasks}
To gain deeper insights into math problems in e-commerce coupon scenarios, we manually analyzed randomly selected examples generated by the best-performing model Deepseek-R1. The categorization results are illustrated in Figure~\ref{fig:error_portion}.
We examined 15 samples that yielded incorrect answers and categorized them accordingly. The examples from each category can be found in Appendix~\ref{app:error_analysis}.

The most prevalent error type is logical mistakes, accounting for 87\% of the errors. These mistakes occur especially when the model is faced with diverse coupon constraints that require extra reasoning capabilities to handle (an example seen in Figure~\ref{fig:part2_logic} in Appendix \ref{app:error_analysis}). 
The second error type is commonsense mistakes, constituting 13\% of the errors, which involve misunderstanding about concepts such as types of goods in specific domains (an example seen in Figure~\ref{fig:commonerror} in Appendix \ref{app:error_analysis}).
The analysis reveals insufficient domain-specific guideline following ability of existing models, especially when handling complex and interdependent multiple guideline rules, which can result in confusion or erroneous outcomes.

\paragraph{Reasoning Strategies and Cognitive Load Analysis}
We highlight two key insights: the reasoning paradigms employed and cognitive load management of LLMs.

(1) Reasoning paradigms. Our experimental results reveal performance variance across reasoning strategies. Inductive pattern recognition proves robust due to its alignment with pretraining data. As shown in Table~\ref{tab:w1}, deductive reasoning strategies, however, exhibit more varied performance:
Specific rule math conversion is highly effective. Manually converting guideline rules into mathematically equivalent expressions, denoted as Equivalent Math, improves accuracy by +21.16\%;
In-context learning techniques, such as employing 2-shot demonstrations and randomly shuffling the order of guideline rules, respectively denoted as 2-shot Demo and Rules reorder, do not contribute to performance enhancement;
PoT~\cite{chen2023programthoughtspromptingdisentangling}, which converts guideline rules into equivalent Python programs via GPT-4o, shows large degradation with the accuracy reduced by 35.11\%.

(2) Cognitive Load. Our analysis also identifies two failure modes related to cognitive load. First, Intrinsic Load Limits: Smaller models (Qwen2.5-7B, Vicuna-7B, and Llama3-8B-Instruct) struggle with tasks requiring simultaneous content compression and output format alignment, suggesting that such tasks impose a cognitive load beyond the capacity of their parameter-efficient attention mechanisms.
Second, Extraneous Load Sensitivity: Removing guideline structure from GPT-4o reduces accuracy by over 5\%, indicating its reliance on structured decomposition to handle non-intrinsic task complexity.

Therefore, we propose two directions two directions to enhance guideline-following capabilities: developing dynamic reasoning modules that adapt thinking depth to task complexity and introducing algorithms such as reinforcement learning to eliminate overlooked rules.

\paragraph{Summary}
Based on experimental results and further analysis, we conclude that guidelines play a crucial role in enhancing instruction following capabilities. While API-based models exhibit strong performance, they are not suitable for operational tasks. In contrast, open-source models still show a significant performance gap. Moreover, CoT offers substantial improvements for complex tasks but provides limited benefits for simpler ones. Identified phenomena of over-thinking and under-thinking introduce new optimization challenges in reasoning efficiency and depth control. Through error analysis, we highlight considerable room for improvement in domain-specific instruction following, particularly in math problems. These findings underscore the challenges posed by complex domains and the need for continued advancements in LLMs’ instruction following abilities.

\section{Conclusion}
\label{sec:conclusion}

LLMs have been increasingly deployed as domain-oriented agents relying on domain-oriented guidelines.
The absence of such benchmarks presents a significant obstacle to effective assessment and further development.  
This paper introduces \BENCH{}, a comprehensive benchmark designed to evaluate guideline following performance of LLM agents. 
\BENCH{} covers 7 distinct categories with 1272 instances.
Our extensive experiments over 18 mainstream LLMs reveal the insufficient instruction following capabilities of current LLMs in complex scenarios, especially for math tasks.
Further in-depth analyses highlight the pivotal role of domain guidelines and indicate substantial opportunities for improving instruction following ability of current LLMs in the future.

\section*{Limitations}
\BENCH{} is primarily constructed based on Chinese instructions, which may neglect some elements in other languages and cultures that can influence the complexity of instructions. Recognizing this constraint, we plan to expand \BENCH{} by incorporating multiple languages to investigate the disparities in complex instruction-following ability of LLMs across different linguistic environments in future iterations.
\section*{Ethics Statement}

\BENCH{} incorporates domain-specific data, and we have taken comprehensive measures to address potential privacy risks throughout the data collection and construction process, ensuring rigorous de-identification and anonymization. In addition, during the manual review phase, we carefully assessed and mitigated any possible societal impacts of the data, adhering to strict ethical guidelines. Furthermore, our work contributes to the advancement of auditing and operational AI agents, which plays an essential role in filtering out ethical risks.

\bibliography{custom}

\begin{thebibliography}{51}
\providecommand{\natexlab}[1]{#1}

\bibitem[{Achiam et~al.(2023)Achiam, Adler, Agarwal, Ahmad, Akkaya, Aleman, Almeida, Altenschmidt, Altman, Anadkat et~al.}]{openai2024gpt4technicalreport}
Josh Achiam, Steven Adler, Sandhini Agarwal, Lama Ahmad, Ilge Akkaya, Florencia~Leoni Aleman, Diogo Almeida, Janko Altenschmidt, Sam Altman, Shyamal Anadkat, et~al. 2023.
\newblock Gpt-4 technical report.
\newblock \emph{arXiv preprint arXiv:2303.08774}.

\bibitem[{Aher et~al.(2023)Aher, Arriaga, and Kalai}]{aher2023using}
Gati~V Aher, Rosa~I Arriaga, and Adam~Tauman Kalai. 2023.
\newblock Using large language models to simulate multiple humans and replicate human subject studies.
\newblock In \emph{International Conference on Machine Learning}, pages 337--371. PMLR.

\bibitem[{Akata et~al.(2023)Akata, Schulz, Coda-Forno, Oh, Bethge, and Schulz}]{akata2023playing}
Elif Akata, Lion Schulz, Julian Coda-Forno, Seong~Joon Oh, Matthias Bethge, and Eric Schulz. 2023.
\newblock \href {https://arxiv.org/abs/2305.16867} {Playing repeated games with large language models}.
\newblock \emph{ArXiv preprint}, abs/2305.16867.

\bibitem[{Boiko et~al.(2023)Boiko, MacKnight, and Gomes}]{boiko2023emergent}
Daniil~A Boiko, Robert MacKnight, and Gabe Gomes. 2023.
\newblock \href {https://arxiv.org/abs/2304.05332} {Emergent autonomous scientific research capabilities of large language models}.
\newblock \emph{ArXiv preprint}, abs/2304.05332.

\bibitem[{Bran et~al.(2023)Bran, Cox, White, and Schwaller}]{bran2023chemcrow}
Andres~M Bran, Sam Cox, Andrew~D White, and Philippe Schwaller. 2023.
\newblock \href {https://arxiv.org/abs/2304.05376} {Chemcrow: Augmenting large-language models with chemistry tools}.
\newblock \emph{ArXiv preprint}, abs/2304.05376.

\bibitem[{Chen et~al.(2023)Chen, Ma, Wang, and Cohen}]{chen2023programthoughtspromptingdisentangling}
Wenhu Chen, Xueguang Ma, Xinyi Wang, and William~W. Cohen. 2023.
\newblock \href {https://arxiv.org/abs/2211.12588} {Program of thoughts prompting: Disentangling computation from reasoning for numerical reasoning tasks}.
\newblock \emph{Preprint}, arXiv:2211.12588.

\bibitem[{Chen et~al.(2025)Chen, Xu, Liang, He, Pang, Yu, Song, Liu, Zhou, Zhang, Wang, Tu, Mi, and Yu}]{chen2025think23overthinkingo1like}
Xingyu Chen, Jiahao Xu, Tian Liang, Zhiwei He, Jianhui Pang, Dian Yu, Linfeng Song, Qiuzhi Liu, Mengfei Zhou, Zhuosheng Zhang, Rui Wang, Zhaopeng Tu, Haitao Mi, and Dong Yu. 2025.
\newblock \href {https://arxiv.org/abs/2412.21187} {Do not think that much for 2+3=? on the overthinking of o1-like llms}.
\newblock \emph{Preprint}, arXiv:2412.21187.

\bibitem[{Chen and Zhang(2024)}]{chen2024revolutionizing}
Xinyu Chen and Lianzhen Zhang. 2024.
\newblock Revolutionizing bridge operation and maintenance with llm-based agents: An overview of applications and insights.
\newblock \emph{arXiv preprint arXiv:2407.10064}.

\bibitem[{Chiang et~al.(2023)Chiang, Li, Lin, Sheng, Wu, Zhang, Zheng, Zhuang, Zhuang, Gonzalez et~al.}]{vicuna2023}
Wei-Lin Chiang, Zhuohan Li, Zi~Lin, Ying Sheng, Zhanghao Wu, Hao Zhang, Lianmin Zheng, Siyuan Zhuang, Yonghao Zhuang, Joseph~E Gonzalez, et~al. 2023.
\newblock Vicuna: An open-source chatbot impressing gpt-4 with 90\%* chatgpt quality.
\newblock \emph{See https://vicuna. lmsys. org (accessed 14 April 2023)}, 2(3):6.

\bibitem[{Dan et~al.(2023)Dan, Lei, Gu, Li, Yin, Lin, Ye, Tie, Zhou, Wang et~al.}]{dan2023educhat}
Yuhao Dan, Zhikai Lei, Yiyang Gu, Yong Li, Jianghao Yin, Jiaju Lin, Linhao Ye, Zhiyan Tie, Yougen Zhou, Yilei Wang, et~al. 2023.
\newblock \href {https://arxiv.org/abs/2308.02773} {Educhat: A large-scale language model-based chatbot system for intelligent education}.
\newblock \emph{ArXiv preprint}, abs/2308.02773.

\bibitem[{Dubois et~al.(2024)Dubois, Liang, and Hashimoto}]{dubois2024lengthcontrolledalpacaevalsimpleway}
Yann Dubois, Percy Liang, and Tatsunori Hashimoto. 2024.
\newblock \href {https://openreview.net/forum?id=CybBmzWBX0} {Length-controlled alpacaeval: A simple debiasing of automatic evaluators}.
\newblock In \emph{First Conference on Language Modeling}.

\bibitem[{Grattafiori et~al.(2024)Grattafiori, Dubey, Jauhri, Pandey, Kadian, Al-Dahle, Letman, Mathur, Schelten et~al.}]{grattafiori2024llama3herdmodels}
Aaron Grattafiori, Abhimanyu Dubey, Abhinav Jauhri, Abhinav Pandey, Abhishek Kadian, Ahmad Al-Dahle, Aiesha Letman, Akhil Mathur, Alan Schelten, et~al. 2024.
\newblock The llama 3 herd of models.
\newblock \emph{arXiv preprint arXiv:2407.21783}.

\bibitem[{He et~al.(2024{\natexlab{a}})He, Zeng, Huang, Chen, Xiao, He, Zhou, Liang, and Xiao}]{he2024can}
Qianyu He, Jie Zeng, Wenhao Huang, Lina Chen, Jin Xiao, Qianxi He, Xunzhe Zhou, Jiaqing Liang, and Yanghua Xiao. 2024{\natexlab{a}}.
\newblock Can large language models understand real-world complex instructions?
\newblock In \emph{Proceedings of the AAAI Conference on Artificial Intelligence}, pages 18188--18196.

\bibitem[{He et~al.(2024{\natexlab{b}})He, Zeng, Huang, Chen, Xiao, He, Zhou, Liang, and Xiao}]{he2024largelanguagemodelsunderstand}
Qianyu He, Jie Zeng, Wenhao Huang, Lina Chen, Jin Xiao, Qianxi He, Xunzhe Zhou, Jiaqing Liang, and Yanghua Xiao. 2024{\natexlab{b}}.
\newblock Can large language models understand real-world complex instructions?
\newblock In \emph{Proceedings of the AAAI Conference on Artificial Intelligence}, pages 18188--18196.

\bibitem[{Hendler(1999)}]{hendler1999there}
James Hendler. 1999.
\newblock Is there an intelligent agent in your future?
\newblock \emph{Nature}, 11.

\bibitem[{Heo et~al.(2024)Heo, Heinze-Deml, Elachqar, Ren, Chan, Nallasamy, Miller, and Narain}]{heollms}
Juyeon Heo, Christina Heinze-Deml, Oussama Elachqar, Shirley~You Ren, Kwan Ho~Ryan Chan, Udhyakumar Nallasamy, Andrew Miller, and Jaya Narain. 2024.
\newblock Do llms internally``know''when they follow instructions?
\newblock In \emph{NeurIPS 2024 Workshop: Foundation Model Interventions}.

\bibitem[{Jiang et~al.(2024)Jiang, Wang, Zeng, Zhong, Li, Mi, Shang, Jiang, Liu, and Wang}]{jiang-etal-2024-followbench}
Yuxin Jiang, Yufei Wang, Xingshan Zeng, Wanjun Zhong, Liangyou Li, Fei Mi, Lifeng Shang, Xin Jiang, Qun Liu, and Wei Wang. 2024.
\newblock \href {https://doi.org/10.18653/v1/2024.acl-long.257} {{F}ollow{B}ench: A multi-level fine-grained constraints following benchmark for large language models}.
\newblock In \emph{Proceedings of the 62nd Annual Meeting of the Association for Computational Linguistics (Volume 1: Long Papers)}, pages 4667--4688, Bangkok, Thailand. Association for Computational Linguistics.

\bibitem[{Kang and Kim(2023)}]{kang2023chatmof}
Yeonghun Kang and Jihan Kim. 2023.
\newblock \href {https://arxiv.org/abs/2308.01423} {Chatmof: An autonomous ai system for predicting and generating metal-organic frameworks}.
\newblock \emph{ArXiv preprint}, abs/2308.01423.

\bibitem[{Kojima et~al.(2023)Kojima, Gu, Reid, Matsuo, and Iwasawa}]{kojima2023largelanguagemodelszeroshot}
Takeshi Kojima, Shixiang~Shane Gu, Machel Reid, Yutaka Matsuo, and Yusuke Iwasawa. 2023.
\newblock \href {https://arxiv.org/abs/2205.11916} {Large language models are zero-shot reasoners}.
\newblock \emph{Preprint}, arXiv:2205.11916.

\bibitem[{Li et~al.(2023{\natexlab{a}})Li, Hammoud, Itani, Khizbullin, and Ghanem}]{li2023camel}
Guohao Li, Hasan Abed Al~Kader Hammoud, Hani Itani, Dmitrii Khizbullin, and Bernard Ghanem. 2023{\natexlab{a}}.
\newblock \href {https://arxiv.org/abs/2303.17760} {Camel: Communicative agents for" mind" exploration of large scale language model society}.
\newblock \emph{ArXiv preprint}, abs/2303.17760.

\bibitem[{Li et~al.(2023{\natexlab{b}})Li, Zhang, Dubois, Taori, Ishaan~Gulrajani, Liang, and Hashimoto}]{lialpacaeval}
Xuechen Li, Tianyi Zhang, Yann Dubois, Rohan Taori, CG~Ishaan~Gulrajani, P~Liang, and TB~Hashimoto. 2023{\natexlab{b}}.
\newblock Alpacaeval: an automatic evaluator of instruction-following models.
\newblock \emph{URL https://github. com/tatsu-lab/alpaca\_eval}.

\bibitem[{Liu et~al.(2024{\natexlab{a}})Liu, Feng, Xue, Wang, Wu, Lu, Zhao, Deng, Zhang, Ruan et~al.}]{deepseekai2024deepseekv3technicalreport}
Aixin Liu, Bei Feng, Bing Xue, Bingxuan Wang, Bochao Wu, Chengda Lu, Chenggang Zhao, Chengqi Deng, Chenyu Zhang, Chong Ruan, et~al. 2024{\natexlab{a}}.
\newblock Deepseek-v3 technical report.
\newblock \emph{arXiv preprint arXiv:2412.19437}.

\bibitem[{Liu et~al.(2024{\natexlab{b}})Liu, Lei, Wang, Huang, Feng, Wen, Cheng, Ke, Xu, Tam, Zhang, Sun, Gu, Wang, Zhang, Huang, Dong, and Tang}]{liu2024alignbenchbenchmarkingchinesealignment}
Xiao Liu, Xuanyu Lei, Shengyuan Wang, Yue Huang, Andrew Feng, Bosi Wen, Jiale Cheng, Pei Ke, Yifan Xu, Weng~Lam Tam, Xiaohan Zhang, Lichao Sun, Xiaotao Gu, Hongning Wang, Jing Zhang, Minlie Huang, Yuxiao Dong, and Jie Tang. 2024{\natexlab{b}}.
\newblock \href {https://doi.org/10.18653/v1/2024.acl-long.624} {{A}lign{B}ench: Benchmarking {C}hinese alignment of large language models}.
\newblock In \emph{Proceedings of the 62nd Annual Meeting of the Association for Computational Linguistics (Volume 1: Long Papers)}, pages 11621--11640, Bangkok, Thailand. Association for Computational Linguistics.

\bibitem[{Ma et~al.(2023)Ma, Mei, and Su}]{ma2023understanding}
Zilin Ma, Yiyang Mei, and Zhaoyuan Su. 2023.
\newblock \href {https://arxiv.org/abs/2307.15810} {Understanding the benefits and challenges of using large language model-based conversational agents for mental well-being support}.
\newblock \emph{ArXiv preprint}, abs/2307.15810.

\bibitem[{Maes(1995)}]{maes1995agents}
Pattie Maes. 1995.
\newblock Agents that reduce work and information overload.
\newblock In \emph{Readings in human--computer interaction}, pages 811--821. Elsevier.

\bibitem[{Mehta et~al.(2023)Mehta, Teruel, Sanz, Deng, Awadallah, and Kiseleva}]{mehta2023improving}
Nikhil Mehta, Milagro Teruel, Patricio~Figueroa Sanz, Xin Deng, Ahmed~Hassan Awadallah, and Julia Kiseleva. 2023.
\newblock \href {https://arxiv.org/abs/2304.10750} {Improving grounded language understanding in a collaborative environment by interacting with agents through help feedback}.
\newblock \emph{ArXiv preprint}, abs/2304.10750.

\bibitem[{Mu et~al.(2023)Mu, Chen, Wang, Chen, Karamardian, Aljeraisy, Hendrycks, and Wagner}]{mu2023can}
Norman Mu, Sarah Chen, Zifan Wang, Sizhe Chen, David Karamardian, Lulwa Aljeraisy, Dan Hendrycks, and David Wagner. 2023.
\newblock Can llms follow simple rules?
\newblock \emph{arXiv preprint arXiv:2311.04235}.

\bibitem[{Ouyang et~al.(2022)Ouyang, Wu, Jiang, Almeida, Wainwright, Mishkin, Zhang, Agarwal, Slama, Ray et~al.}]{ouyang2022training}
Long Ouyang, Jeffrey Wu, Xu~Jiang, Diogo Almeida, Carroll Wainwright, Pamela Mishkin, Chong Zhang, Sandhini Agarwal, Katarina Slama, Alex Ray, et~al. 2022.
\newblock Training language models to follow instructions with human feedback.
\newblock \emph{Advances in neural information processing systems}, 35:27730--27744.

\bibitem[{Qian et~al.(2023)Qian, Cong, Yang, Chen, Su, Xu, Liu, and Sun}]{qian2023communicative}
Chen Qian, Xin Cong, Cheng Yang, Weize Chen, Yusheng Su, Juyuan Xu, Zhiyuan Liu, and Maosong Sun. 2023.
\newblock \href {https://arxiv.org/abs/2307.07924} {Communicative agents for software development}.
\newblock \emph{ArXiv preprint}, abs/2307.07924.

\bibitem[{Qin et~al.(2024{\natexlab{a}})Qin, Song, Hu, Yao, Cho, Wang, Wu, Liu, Liu, and Yu}]{qin-etal-2024-infobench}
Yiwei Qin, Kaiqiang Song, Yebowen Hu, Wenlin Yao, Sangwoo Cho, Xiaoyang Wang, Xuansheng Wu, Fei Liu, Pengfei Liu, and Dong Yu. 2024{\natexlab{a}}.
\newblock \href {https://doi.org/10.18653/v1/2024.findings-acl.772} {{I}n{F}o{B}ench: Evaluating instruction following ability in large language models}.
\newblock In \emph{Findings of the Association for Computational Linguistics: ACL 2024}, pages 13025--13048, Bangkok, Thailand. Association for Computational Linguistics.

\bibitem[{Qin et~al.(2024{\natexlab{b}})Qin, Song, Hu, Yao, Cho, Wang, Wu, Liu, Liu, and Yu}]{qin2024infobenchevaluatinginstructionfollowing}
Yiwei Qin, Kaiqiang Song, Yebowen Hu, Wenlin Yao, Sangwoo Cho, Xiaoyang Wang, Xuansheng Wu, Fei Liu, Pengfei Liu, and Dong Yu. 2024{\natexlab{b}}.
\newblock \href {https://doi.org/10.18653/v1/2024.findings-acl.772} {{I}n{F}o{B}ench: Evaluating instruction following ability in large language models}.
\newblock In \emph{Findings of the Association for Computational Linguistics: ACL 2024}, pages 13025--13048, Bangkok, Thailand. Association for Computational Linguistics.

\bibitem[{Roy et~al.(2024)Roy, Zhang, Bhave, Bansal, Las-Casas, Fonseca, and Rajmohan}]{roy2024exploring}
Devjeet Roy, Xuchao Zhang, Rashi Bhave, Chetan Bansal, Pedro Las-Casas, Rodrigo Fonseca, and Saravan Rajmohan. 2024.
\newblock Exploring llm-based agents for root cause analysis.
\newblock In \emph{Companion Proceedings of the 32nd ACM International Conference on the Foundations of Software Engineering}, pages 208--219.

\bibitem[{Searle(1969)}]{searle1969speech}
John~R Searle. 1969.
\newblock \emph{Speech acts: An essay in the philosophy of language}, volume 626.
\newblock Cambridge university press.

\bibitem[{Sun et~al.(2024)Sun, Zhang, Zhang, Yu, Huang, Chen, Xu, He, Zhao, and Liu}]{sun2024beyond}
Wangtao Sun, Chenxiang Zhang, XueYou Zhang, Xuanqing Yu, Ziyang Huang, Pei Chen, Haotian Xu, Shizhu He, Jun Zhao, and Kang Liu. 2024.
\newblock Beyond instruction following: Evaluating inferential rule following of large language models.
\newblock \emph{arXiv preprint arXiv:2407.08440}.

\bibitem[{Tang et~al.(2024)Tang, Zong, Phang, Zhao, Zhou, Cohan, and Gerstein}]{tang2024strucbenchlargelanguagemodels}
Xiangru Tang, Yiming Zong, Jason Phang, Yilun Zhao, Wangchunshu Zhou, Arman Cohan, and Mark Gerstein. 2024.
\newblock \href {https://doi.org/10.18653/v1/2024.naacl-short.2} {Struc-bench: Are large language models good at generating complex structured tabular data?}
\newblock In \emph{Proceedings of the 2024 Conference of the North American Chapter of the Association for Computational Linguistics: Human Language Technologies (Volume 2: Short Papers)}, pages 12--34, Mexico City, Mexico. Association for Computational Linguistics.

\bibitem[{Wang et~al.(2023)Wang, Liu, Yue, Tang, Zhang, Jiayang, Yao, Gao, Hu, Qi et~al.}]{wang2023survey}
Cunxiang Wang, Xiaoze Liu, Yuanhao Yue, Xiangru Tang, Tianhang Zhang, Cheng Jiayang, Yunzhi Yao, Wenyang Gao, Xuming Hu, Zehan Qi, et~al. 2023.
\newblock \href {https://arxiv.org/abs/2310.07521} {Survey on factuality in large language models: Knowledge, retrieval and domain-specificity}.
\newblock \emph{ArXiv preprint}, abs/2310.07521.

\bibitem[{Wang et~al.(2025)Wang, Liu, Xu, Liang, Chen, He, Song, Yu, Li, Zhang, Wang, Tu, Mi, and Yu}]{wang2025thoughtsplaceunderthinkingo1like}
Yue Wang, Qiuzhi Liu, Jiahao Xu, Tian Liang, Xingyu Chen, Zhiwei He, Linfeng Song, Dian Yu, Juntao Li, Zhuosheng Zhang, Rui Wang, Zhaopeng Tu, Haitao Mi, and Dong Yu. 2025.
\newblock \href {https://arxiv.org/abs/2501.18585} {Thoughts are all over the place: On the underthinking of o1-like llms}.
\newblock \emph{Preprint}, arXiv:2501.18585.

\bibitem[{Wen et~al.(2024)Wen, Ke, Gu, Wu, Huang, Zhou, Li, Hu, Gao, Xu et~al.}]{wen2024complexbench}
Bosi Wen, Pei Ke, Xiaotao Gu, Lindong Wu, Hao Huang, Jinfeng Zhou, Wenchuang Li, Binxin Hu, Wendy Gao, Jiaxing Xu, et~al. 2024.
\newblock Benchmarking complex instruction-following with multiple constraints composition.
\newblock In \emph{The Thirty-eight Conference on Neural Information Processing Systems Datasets and Benchmarks Track}.

\bibitem[{Xi et~al.(2023)Xi, Chen, Guo, He, Ding, Hong, Zhang, Wang, Jin, Zhou et~al.}]{xi2023rise}
Zhiheng Xi, Wenxiang Chen, Xin Guo, Wei He, Yiwen Ding, Boyang Hong, Ming Zhang, Junzhe Wang, Senjie Jin, Enyu Zhou, et~al. 2023.
\newblock \href {https://arxiv.org/abs/2309.07864} {The rise and potential of large language model based agents: A survey}.
\newblock \emph{ArXiv preprint}, abs/2309.07864.

\bibitem[{Xia et~al.(2024)Xia, Xing, Du, Yang, Feng, Xu, Yin, and Xiong}]{xia2024fofobenchmarkevaluatellms}
Congying Xia, Chen Xing, Jiangshu Du, Xinyi Yang, Yihao Feng, Ran Xu, Wenpeng Yin, and Caiming Xiong. 2024.
\newblock \href {https://doi.org/10.18653/v1/2024.acl-long.40} {{FOFO}: A benchmark to evaluate {LLM}s' format-following capability}.
\newblock In \emph{Proceedings of the 62nd Annual Meeting of the Association for Computational Linguistics (Volume 1: Long Papers)}, pages 680--699, Bangkok, Thailand. Association for Computational Linguistics.

\bibitem[{Xie et~al.(2024)Xie, Zhang, Chen, Lou, and Su}]{xieadaptive}
Jian Xie, Kai Zhang, Jiangjie Chen, Renze Lou, and Yu~Su. 2024.
\newblock Adaptive chameleon or stubborn sloth: Revealing the behavior of large language models in knowledge conflicts.
\newblock In \emph{The Twelfth International Conference on Learning Representations}.

\bibitem[{Xu et~al.(2024{\natexlab{a}})Xu, Sun, Zheng, Geng, Zhao, Feng, Tao, Lin, and Jiang}]{xu2023wizardlmempoweringlargelanguage}
Can Xu, Qingfeng Sun, Kai Zheng, Xiubo Geng, Pu~Zhao, Jiazhan Feng, Chongyang Tao, Qingwei Lin, and Daxin Jiang. 2024{\natexlab{a}}.
\newblock \href {https://openreview.net/forum?id=CfXh93NDgH} {Wizard{LM}: Empowering large pre-trained language models to follow complex instructions}.
\newblock In \emph{The Twelfth International Conference on Learning Representations}.

\bibitem[{Xu et~al.(2024{\natexlab{b}})Xu, Qi, Guo, Wang, Wang, Zhang, and Xu}]{xu-etal-2024-knowledge-conflicts}
Rongwu Xu, Zehan Qi, Zhijiang Guo, Cunxiang Wang, Hongru Wang, Yue Zhang, and Wei Xu. 2024{\natexlab{b}}.
\newblock \href {https://doi.org/10.18653/v1/2024.emnlp-main.486} {Knowledge conflicts for {LLM}s: A survey}.
\newblock In \emph{Proceedings of the 2024 Conference on Empirical Methods in Natural Language Processing}, pages 8541--8565, Miami, Florida, USA. Association for Computational Linguistics.

\bibitem[{Yang et~al.(2024)Yang, Yang, Zhang, Hui, Zheng, Yu, Li, Liu, Huang, Wei et~al.}]{qwen2.5}
An~Yang, Baosong Yang, Beichen Zhang, Binyuan Hui, Bo~Zheng, Bowen Yu, Chengyuan Li, Dayiheng Liu, Fei Huang, Haoran Wei, et~al. 2024.
\newblock Qwen2.5 technical report.
\newblock \emph{arXiv preprint arXiv:2412.15115}.

\bibitem[{Yin et~al.(2023)Yin, Ye, Liu, Ren, and Sch{\"u}tze}]{yin2023llm}
Wenpeng Yin, Qinyuan Ye, Pengfei Liu, Xiang Ren, and Hinrich Sch{\"u}tze. 2023.
\newblock Llm-driven instruction following: Progresses and concerns.
\newblock In \emph{Proceedings of the 2023 Conference on Empirical Methods in Natural Language Processing: Tutorial Abstracts}, pages 19--25.

\bibitem[{Yuan et~al.(2024)Yuan, He, Dong, Wang, Zhao, Xia, Xu, Zhou, Li, Zhang, Wang, and Liu}]{yuan2024rjudgebenchmarkingsafetyrisk}
Tongxin Yuan, Zhiwei He, Lingzhong Dong, Yiming Wang, Ruijie Zhao, Tian Xia, Lizhen Xu, Binglin Zhou, Fangqi Li, Zhuosheng Zhang, Rui Wang, and Gongshen Liu. 2024.
\newblock \href {https://arxiv.org/abs/2401.10019} {R-judge: Benchmarking safety risk awareness for llm agents}.
\newblock \emph{Preprint}, arXiv:2401.10019.

\bibitem[{Zhang et~al.(2023)Zhang, Yao, Zhang, Tang, Ma, He, Wang, Gerstein, Wang, Liu, and Zhao}]{zhang2023ignitinglanguageintelligencehitchhikers}
Zhuosheng Zhang, Yao Yao, Aston Zhang, Xiangru Tang, Xinbei Ma, Zhiwei He, Yiming Wang, Mark Gerstein, Rui Wang, Gongshen Liu, and Hai Zhao. 2023.
\newblock \href {https://arxiv.org/abs/2311.11797} {Igniting language intelligence: The hitchhiker's guide from chain-of-thought reasoning to language agents}.
\newblock \emph{Preprint}, arXiv:2311.11797.

\bibitem[{Zheng et~al.(2023{\natexlab{a}})Zheng, Chiang, Sheng, Zhuang, Wu, Zhuang, Lin, Li, Li, Xing, Zhang, Gonzalez, and Stoica}]{zheng2023judgingllmasajudgemtbenchchatbot}
Lianmin Zheng, Wei-Lin Chiang, Ying Sheng, Siyuan Zhuang, Zhanghao Wu, Yonghao Zhuang, Zi~Lin, Zhuohan Li, Dacheng Li, Eric Xing, Hao Zhang, Joseph~E. Gonzalez, and Ion Stoica. 2023{\natexlab{a}}.
\newblock \href {https://openreview.net/forum?id=uccHPGDlao} {Judging {LLM}-as-a-judge with {MT}-bench and chatbot arena}.
\newblock In \emph{Thirty-seventh Conference on Neural Information Processing Systems Datasets and Benchmarks Track}.

\bibitem[{Zheng et~al.(2023{\natexlab{b}})Zheng, Chiang, Sheng, Zhuang, Wu, Zhuang, Lin, Li, Li, Xing et~al.}]{zheng2023judging}
Lianmin Zheng, Wei-Lin Chiang, Ying Sheng, Siyuan Zhuang, Zhanghao Wu, Yonghao Zhuang, Zi~Lin, Zhuohan Li, Dacheng Li, Eric Xing, et~al. 2023{\natexlab{b}}.
\newblock Judging llm-as-a-judge with mt-bench and chatbot arena.
\newblock \emph{Advances in Neural Information Processing Systems}, 36:46595--46623.

\bibitem[{Zhou et~al.(2023{\natexlab{a}})Zhou, Lu, Mishra, Brahma, Basu, Luan, Zhou, and Hou}]{zhou2023instructionfollowingevaluationlargelanguage}
Jeffrey Zhou, Tianjian Lu, Swaroop Mishra, Siddhartha Brahma, Sujoy Basu, Yi~Luan, Denny Zhou, and Le~Hou. 2023{\natexlab{a}}.
\newblock Instruction-following evaluation for large language models.
\newblock \emph{arXiv preprint arXiv:2311.07911}.

\bibitem[{Zhou et~al.(2023{\natexlab{b}})Zhou, Jiang, Li, Wu, Wang, Qiu, Zhang, Chen, Wu, Wang et~al.}]{zhou2023agents}
Wangchunshu Zhou, Yuchen~Eleanor Jiang, Long Li, Jialong Wu, Tiannan Wang, Shi Qiu, Jintian Zhang, Jing Chen, Ruipu Wu, Shuai Wang, et~al. 2023{\natexlab{b}}.
\newblock \href {https://arxiv.org/abs/2309.07870} {Agents: An open-source framework for autonomous language agents}.
\newblock \emph{ArXiv preprint}, abs/2309.07870.

\end{thebibliography}

\clearpage
\newpage 
\onecolumn

\appendix

\section{Experiment Settings}

\subsection{Hyperparameter}\label{app:Experiment_Settings}
For paramaters, we set temperature=0 and top\_p=1, and use the default values of official model releases for other paramaters.

\subsection{Experiment Cost}
In the case of API-based models, the time required per experiment (traversing all samples in one pass) is generally under 1 hour, although it may be influenced by the state of the network.
The time consumption for each open-sourced model in each experiment is approximately 0.5 hour.

\subsection{Model Information}\label{app:Model_Information}
Table \ref{tab:LLMs_evaluated} lists concrete information about models in the experiments.
\begin{table*}[htb]
  \centering
  \renewcommand\tabcolsep{4pt} 
  \scalebox{0.8}{
  \begin{tabular}{l|cccc}
  \toprule
  \multirow{1}{*}{Models}  
    & Model Size & Access & Version & Creator \\\midrule

    o1  & undisclosed  & api  & o1-2024-12-17 & \multirow{2}{*}{OpenAI}   \\
    GPT-4o & undisclosed & api  & gpt-4o-2024-11-20  &   
     \\\midrule
    
    Deepseek-R1  & undisclosed  & api  & deepseek-r1-2025-01-20 & \multirow{2}{*}{DeepSeek} \\
    Deepseek-V3 & 671B & api  & deepseek-r1-2024-12-26  &  \\ \midrule

    Gemini2.5-pro-exp & undisclosed & api & gemini-2.5-pro-exp-0325 & Google \\ \midrule
    Mistral-7B-Instruct & 7B & weights & -  &Mistral AI \\ \midrule

    Yi-1.5-6B & 6B & weights & -  & \multirow{2}{*}{01.AI } \\
    Yi-1.5-34B & 34B & weights & -  & \\ \midrule

    Gemma-3-4b-it & 4B & weights & - &  Google \\ \midrule
    
    QwQ-32B-Preview & 32B & weights & - & \multirow{6}{*}{Qwen} \\
    
    R1-Distill-Qwen-7B  & 7B & weights & - &   \\
    
    R1-Distill-Qwen-32B  & 32B & weights & -  &  \\
    
    Qwen2.5-7B  & 7B  & weights  & -  &   \\
    Qwen2.5-7B-Instruct  & 7B & weights & - &  \\

    Qwen2.5-32B  & 32B  & weights  & - &  \\ \midrule
    Vicuna-7B-v1.5  & 7B  & weights  & - & LMSYS \\ \midrule

    Llama3-8B-Instruct  & 8B & weights  & -  & \multirow{2}{*}{Meta}  \\
    Llama3.3-70B-Instruct  & 70B & weights  & -  & \\
    \bottomrule
  \end{tabular}}
  \caption{LLMs evaluated in this paper.}
  \label{tab:LLMs_evaluated}
\end{table*}

   
    

    



\section{Detailed Prompts}
\label{app:all_prompts}

\subsection{System Prompts of different task domains}\label{app:grg}
For different domains, we have developed task guidelines and some requirements for rule writing, as shown in Figure \ref{fig:prompt-2} \textendash{} \ref{fig:prompt-6}.

\begin{figure}[!ht]
\begin{lstlisting}[style=prompt]
Compare two products to see if they meet the definition of the same product based on their text descriptions. The input is a text description of a product, and the output result can only be the same or different.\nTo compare whether the products are the same, you need to compare the various attributes of the products. If one of the attributes of the two products is judged to be inconsistent, the two products are considered inconsistent. They are considered the same only when they are exactly the same.\nPlease output strictly in json format.

Rule set writing requirements: 1. Only consider the text attributes of the product itself.\n2. The conditions for attribute comparison are only based on objective conditions such as same, different, one product mentioned and the other not mentioned.\n3. Each product attribute must generate a rule that ignores its difference, and it is forbidden to mention exemption conditions such as "unless" in the rule.\n4. For the case where one product mentions and the other does not mention, there are two versions of rules, one for judging the attribute to be the same and the other for judging the attribute to be different.
\end{lstlisting}
\caption{System prompt for price matching task.}
\label{fig:prompt-2}
\end{figure}

\begin{figure}[!ht]
\begin{lstlisting}[style=prompt]
Determine the relevance of Large Language Model (LLM) Retrieval Augmentation Generation (RAG), the input is the topic of the question and the retrieved information, and the output is the relevance determination result (strong correlation, weak correlation, irrelevant). 
### Strong relevance rules 
1. **High subject fit** 
- **Core content match**: If the retrieved information is highly consistent with the topic of the question in the core concept.
- **High keyword overlap**: The search content contains key professional terms, subject objects and other keywords in the question, and the appearance of these keywords is not accidental and isolated, but is presented in an orderly manner around related topics.  
2. **Can directly answer questions** 
- **Provide clear answers or solutions**: When the search content can directly respond to the questions in the question, and give clear and specific answers, explanations, operation methods, etc., it is strongly relevant.  
- **Targeted supplementary information**: Targeted expansion and in-depth analysis of the topics involved in the question, such as asking "The character characteristics of the protagonist in a certain novel", the search content deeply interprets the reasons for the formation of the protagonist's personality from different plots and events, and the performance of the personality at different stages of the story, etc., which is highly related to the question and is determined to be strongly relevant. 
3. **Logical coherence and adaptation** 
- **Contextual logic is consistent**: The retrieved content is reasonable and smooth within the context and logical framework set by the question, and there will be no logical abruptness or contradiction.  
- **Meet the question restriction conditions**: If there are specific restrictions in the question, such as time range ("Development trend of the e-commerce industry in 2023"), geographical range ("New energy policy in Europe"), object range ("Common faults and solutions of a certain brand and model of mobile phone"), etc., the search content accurately matches these restrictions, and the relevance is strong. 

### Weak correlation rules 
1. **Partial correlation of the topic** 
- **Indirectly related to the topic**: The search content is only related to the question topic to a certain extent, but does not focus on the core issue. 
- **Partial keyword match**: Only some of the keywords in the question are included, or the keywords appear but the semantic association is not tight. 
2. **Auxiliary correlation** 
- **Provide broad background information**: The search content is more about providing broad content such as the general background and basic knowledge related to the question topic. It cannot directly answer the core of the question, but it has a certain role in assisting understanding. 
- **Related concepts are expanded but deviate from the focus**: Some concepts related to the question topic are expanded and introduced, but deviate from the focus of the current question.  
### Irrelevant rules 
1. **Completely irrelevant topics** 
- **Different fields**: The topics of the retrieved content and the topic of the question belong to completely different fields and are completely unrelated.  
- **Different subject objects**: The core object around is very different from the object targeted by the question. 
2. **Unable to provide effective information** 
- **Empty or worthless content**: The retrieved information is either just a pile of sentences without any substantive meaning, or it is of no help in answering the question, such as all slogan-like, general and non-targeted words.  
- **Misleading or illogical**: The content has obvious errors, which will mislead the answer to the question, or the logic is confusing and cannot establish a reasonable connection with the question. 
\end{lstlisting}
\caption{System prompt for text relevance task.}
\label{fig:prompt-3}
\end{figure}

\begin{figure}[!ht]
\begin{lstlisting}[style=prompt]
You need to use coupons from an online platform to buy goods. The types of coupons include coupons for full discounts, discount coupons, and fixed amount coupons. 

Requirements for rule writing: The rules need to include rules for the specific amount, applicable time, applicable scope, and combined use of a type of coupon. The coupon amount needs to be changed. 

The following is an example of a rule: 
Full discount coupon: 50 off for purchases over 200, can be used in combination with other coupons, but can only be used once per order, and is applicable to all products on the platform. \nDiscount coupon: 20% off coupon, applicable to all clothing and footwear products except backpacks, and can only be used once per order. \nFixed amount coupon: 10 yuan coupon, no consumption threshold, can be used to purchase any product, but can only be used once per order.
\end{lstlisting}
\caption{System prompt for math task.}
\label{fig:prompt-6}
\end{figure}

\begin{figure}[!ht]
\begin{lstlisting}[style=prompt]
In the airline flight booking customer service scenario, based on the order information and the conversation content, first determine the user's situation and goal, and then generate a reply discourse requirement that is useful to the user. \n Input includes: order information (whether the flight has been booked), available flight information data, and conversation content (may include user needs). \n

Rule requirements: When the goal is "booking", please determine the customer's travel restrictions and search for available flights in the flight information data. There may be available flights or there may not be. When the goal is "rebooking", the order information should be checked to determine whether there is a booking. If there is, interact with the customer to determine the travel restrictions. Otherwise, the status operation "no booking" will be selected, and then search for available flights in the flight information data. There may be available flights or there may not be. Finally, for customers who want to cancel the ticket, the agent will perform a simple check. If a booking is found, it will be cancelled. Otherwise, the conversation will conclude that "no booking". Please determine the action that should be taken for each situation and give the corresponding response. \n

The response may include the following situations: \n1. Ask the user more detailed questions or demands, such as "Did you encounter any problems?" "Do you want to ask xx question?" This kind of situation. \n2. Generate reasonable words to reply to users according to the conditions, respond to users' demands (such as refunds, compensation, complaints, etc.), and provide solutions. \n3. Comfort users when they are in a negative mood. \n4. Use "I'm sorry" as the beginning to indicate that you cannot reply. \n5. Further confirm key information, guide customers to think about related issues, or verify whether the solution meets customer needs. \n

Specific requirements for answer output: \nPlease output the reply to the user based on the given order information and conversation content. \nRule writing requirements: The rules only give the user's goals and reply requirements, and do not need to give specific reply content.
\end{lstlisting}
\caption{System prompt for agent chatting task.}
\label{fig:prompt-7}
\end{figure}

\begin{figure}[!ht]
\begin{lstlisting}[style=prompt]
I need to generate summary requirements for general text. First think about what content/elements are in the text, then selectively include some content and not include some content. The input is general text, and the output is a summary of the text according to the rules. \n

Rule examples: "If the text involves the background of the event, it needs to be summarized in the summary, but there is no need to mention the results of the event. At the same time, it is necessary to ensure that the data content in the text remains the same without modification, and contains three paragraphs." "When summarizing, it is necessary to summarize the numerical information in the text, and there is no need to mention the location information in the text. It should not exceed 150-200 words.". \n

Rule set writing requirements: 
1. The rules need to have requirements for "need to include" or "no need to mention" in the content, and clarify whether various types of content, elements, and attributes need to be summarized or the original text needs to be kept unchanged. 
2. Each type of attribute must have two rules of "need to include" or "no need to mention". If the attributes are the same, just change the verb directly, without adding adjectives, and it does not need to be consistent with the conventions of regular summaries. 
3. The verbs in the rules are only allowed to use clear and objectively verifiable actions, such as "include", "mention", "no need to mention", "keep the original text unchanged", etc. 
4. There need to be requirements on the summary output format, such as language, number of words, number of paragraphs, number of sentences, etc.
\end{lstlisting}
\caption{System prompt for summarization task.}
\label{fig:prompt-4}
\end{figure}

\begin{figure}[!ht]
\begin{lstlisting}[style=prompt]
You are required to give a judgment on whether the <generated content> has hallucination problems and the classification of hallucinations. If you think that the <generated content> has hallucination problems, you are required to give the major categories and subcategories of hallucination problems. \n

The major categories and subcategories are as follows: \n
1. Factual hallucinations\na. Fabrication of facts: making up false facts out of nothing. \nb. Exaggeration or reduction of facts: The generated content contains exaggerated\nor reduced facts. \n
2. Logical hallucinations\na. Causal errors: incorrectly inferring causal relationships, although the two may only be related or completely unrelated. \nb. Logical deduction errors: there are wrong steps in the reasoning process, resulting in conclusions that do not match the premise. \nc. Number, proportion, time errors: errors in calculations related to quantitative relationships, proportions or time. \nd. Inductive and deductive errors: drawing overly broad or incorrect conclusions from specific data, or inferring conclusions that contradict specific examples from general rules. \n
3. Semantic hallucinations\na. Semantic contradictions: The generated content is semantically contradictory, and the statements before and after are conflicting. \nb. Out of context: The generated content fails to keep pace with the context or deviates from the question. \n c. Inappropriate use of words: The words used do not fit the context or are inappropriate and irregular. \n 

The reply should indicate whether hallucinations occur: If there are no hallucinations, just output [No hallucinations]; if there are, indicate the major and subdivided hallucination categories.
\end{lstlisting}
\caption{System prompt for hallucination detection task.}
\label{fig:prompt-5}
\end{figure}

\clearpage 
\FloatBarrier
\subsection{Prompt Template for Generating Rules}\label{app:ptgr}
Figure \ref{fig:prompt-1} is the prompt for extracting key information using large language models. Figure \ref{fig:prompt-8} is the guideline for writing the rule set according to the task instructions. Figure \ref{fig:prompt-9} is the reference for rewriting rules to generate different outputs for the same input.
Figure \ref{fig:prompt-10} is for classifying rules and adding new categories.

\begin{figure}[!ht]
\begin{lstlisting}[style=prompt]
- Please analyze and process the following task guidelines, extracting key information as specified while preserving the original text for other content.

- The key information to be extracted includes:
  1. **objective**: Summarize the overall purpose of the task in a single sentence.
  2. **input**: Identify and list the required input elements for the task as mentioned in the text.
  3. **output**: Describe the expected output of the task, including both content and format.
  4. **requirements**: Specify any requirements for the rule set; if none are mentioned, return an empty string.

- Other content should be preserved:
  **other_content**: Retain the original text in full, excluding only the formatting of the output. Ensure that the remaining content remains unaltered without summarization or omission.

- Please return the results in JSON format as follows: 
{
    "objective": "Briefly describe the mission objectives",
    "input": "Expected input",
    "output": {
        "content": "Expected output",
        "format": "Describes the specific format of the output"
    },
    "requirements": "",
    "other_content": "Keep other original content intact"
}.
\end{lstlisting}
\caption{Prompt template for extracting key information from the guideline.}
\label{fig:prompt-1}
\end{figure}

\begin{figure}[!ht]
\begin{lstlisting}[style=prompt]
Please formulate a rule set for the specified task based on the provided task guidelines. A rule set consists of multiple specific rules and should be returned according to the specified requirements:

1. **Requirements for drafting the rules:**  
   {self.requirements}

2. **Rule set requirements:**  
   - The rule set must comprehensively and thoroughly cover all situations mentioned in the task guidelines, ensuring no omissions.  
   - The rule set should not contain any specially defined terms; replace such terms with their corresponding explanations.  

3. **Output format requirements:**  
   - The final rule set must be returned in JSON format, following the structure: {json_format}.  
   - Each field in the JSON represents a specific rule.
\end{lstlisting}
\caption{Guideline prompt for generating rules.}
\label{fig:prompt-8}
\end{figure}

\begin{figure}[!ht]
\begin{lstlisting}[style=prompt]
There is an existing rule set for the **{self.target}** task. The rule set consists of a collection of specific regulations and is used to evaluate input **{self.input_content}** and generate the corresponding output **{self.output_content}**.  

To enhance the system's flexibility and adaptability, different rules can be selected based on various requirements to accommodate multiple application scenarios. Therefore, please analyze the existing task rules and modify the **evaluation criteria and applicable scope** to generate new rules, ensuring that the output changes under the same input conditions.

**Requirements for drafting the rules:**  
{self.requirements}  

Return only the newly generated rules in **JSON format**, with no explanations or additional information. The format should follow: {json_format}.
\end{lstlisting}
\caption{Guideline prompt for editing rules.}
\label{fig:prompt-9}
\end{figure}

\begin{figure*}[!ht]
\begin{lstlisting}[style=prompt]
I need to categorize and organize a given rule set. The rule set consists of a collection of specific rules. Please return the results in **JSON format**.  

## Requirements:  
- Create a clear and concise classification structure and assign an appropriate name to each category. Multi-level categorization is allowed.  
- If a rule is related to multiple categories, assign it to the most suitable single category. Ensure that each rule belongs to only **one** category.  
- Retain the original and complete content of each rule.  

**JSON Output Format:**  
- Each JSON object should contain **two keys**:  
  - `"type"`: Represents the category name.  
  - `"atomic_rules"`: Contains all rules that belong to the specified category.  
- **Return only JSON format** without any explanations or additional information. The format should follow: {json_format}.
\end{lstlisting}
\caption{Guideline prompt for categorizing rules.}
\label{fig:prompt-10}
\end{figure*}

\begin{figure*}[!ht]
\begin{lstlisting}[style=prompt]
There is an existing rule set for the **{self.target}** task. The rule set consists of a collection of specific rules used to evaluate input **{self.input_content}** and generate the corresponding output **{self.output_content}**.  

Please analyze whether the current rule classification is comprehensive. If any categories are missing, add new categories along with relevant rules.  

The newly added categories and rules should be incorporated into the existing rule set and returned in **JSON format**.  

**Requirements for drafting the rules:**  
{self.requirements}  

**Return format:**  
- The JSON object should contain **two keys**: `"type"` and `"atomic_rules"`.  
- `"type"`: A string representing the category name.  
- `"atomic_rules"`: A nested JSON object containing a set of atomic rules. Inside `"atomic_rules"`, each key should follow the format `"rule_" + number, representing individual rules.  

**Return only JSON format** without any explanations or additional information. The format should follow: {json_format}.
\end{lstlisting}
\caption{Guideline prompt for adding rules.}
\label{fig:prompt-11}
\end{figure*}

\FloatBarrier
\subsection{Prompt for Eliminating Duplicate Rules}\label{duplicate}
We use GPT-4o to remove redundant rules from the rule set. The prompt is as Figure \ref{fig:duplicate-1}.
\begin{figure*}[!ht]
\begin{lstlisting}[style=prompt]
Please deduplicate the given rule set as required and keep the original JSON format.

Complete redundancy: The conditions and outputs of a rule are exactly the same as those of another rule and can be deleted directly.
Implicit redundancy: The effect of a rule is completely covered by another more comprehensive rule and does not need to exist separately.
Conflicting redundancy: The conditions of two rules are similar but the outputs are the same or contradictory, and need to be merged or resolved.
Invalid redundancy: The rule does not contribute to the final decision (such as it is impossible to trigger or blocked by a higher-level rule).
\end{lstlisting}
\caption{Prompt for eliminating duplicate rules.}
\label{fig:duplicate-1}
\end{figure*}

\subsection{Prompts for Generating Context}\label{app:gc}
The following are prompts for generating context for different types of tasks, as shown in Figure \ref{fig:context-1} \textendash{} \ref{fig:context-6}.

\FloatBarrier

\begin{figure}[!ht]
\begin{lstlisting}[style=prompt]
Compare two products to see if they meet the definition of the same product based on their text descriptions":"""You are a question-setting assistant and need to complete question-setting tasks based on different goals. These questions will be used to evaluate the instruction-following ability of the large language model.

First, the objectives of the task are as follows:
<objective>{obj}</objective>

And, you need to follow the following {upper} rules when judging:
<rules>{rules}</rules>

The generated questions are as follows, avoiding too much repetition:
<questions>{question}</questions>

When setting questions, you need to meet the following output requirements:
1. Give different generated content for the two cases where the judgment results are "same product" and "not the same product".
2. The generated content should include product titles and product details. The titles and details of products A and B with the judgment results of "same product" should not be too similar.
3. The generated content should conform to the e-commerce copywriting style. Do not output "a certain brand", nor directly output "the sales channel is xxx", "the ingredients are xxx", do not say "xx is not mentioned", and do not ask "whether it meets the definition of the same product" at the end.
4. Carefully select rules to construct errors, and the errors should not be too obvious. The generated content must meet at least {lower} rules.
5. Multiple questions are output in a list, and each question is output in json format. Each question contains 4 fields, namely "question", "label", "analysis" and "picked_rules". The content of each question in the "question" field is a string, and product A and product B must include product titles and product details, but no result analysis. The "label" field has only 2 results, 0 represents the final judgment of "not the same product" question, and 1 represents the final judgment of "the same product" question. The "analysis" field is a string, which is the reason analysis for judging <objective>. Do not mention the rule_id based on the rule, just say the content directly. The "picked_rules" field represents the rules selected for generating content. It is a JSON list containing {lower} rules, retaining the "type", "rule_id", and "rule_text" of the selected rule.

Now, please make questions according to the above requirements.
\end{lstlisting}
\caption{Prompt for price matching.}
\label{fig:context-1}
\end{figure}

\begin{figure}[!ht]
\begin{lstlisting}[style=prompt]
You are a question-setting assistant. Your task is to set questions based on specific goals and rules to evaluate the instruction-following ability of the large language model.
First, the goal of the task:
<objective>{obj}</objective>

And, you need to follow the following {upper} rules when judging:
<rules>{rules}</rules>

The generated questions are as follows, avoiding too much repetition:
<questions>{question}</questions>

Now you need to set 3 questions based on <rules> to judge the relevance of RAG content and user query, and judge "strongly relevant", "weakly relevant", and "irrelevant" respectively (according to the definition in <rules>)

The requirements are as follows:
- In e-commerce or customer service scenarios, the generated content must include user questions [query] and text segments [doc] obtained by the model RAG. User questions are oral, while RAG text is more written.
- Both [query] and [doc] need to be specific and diversified, including at least 25 characters, including complex sentence structures.
- [query] and [doc] must be able to judge "strong correlation" or "weak correlation" or "irrelevant" (according to the definition in <rules>)

The questions need to be output in a list in json format, and each question contains 3 fields:

1. "question" field: fill in the copy content here, in string form, excluding any results and analysis.

2. "label" field: there are only 3 results, 2 represents the question of strong correlation, 1 represents the question of weak correlation, and 0 represents the question of irrelevant.

3. "analysis" field: get a brief reason for the judgment result, just explain the content directly, and do not mention the rule_id based on the rule.

4. "picked_rules" field: a JSON list containing {lower} rules, retaining the "type", "rule_id", and "rule_text" of the selected rule.

Now please ask questions according to the requirements.
\end{lstlisting}
\caption{Prompt for text relevance.}
\label{fig:context-5}
\end{figure}

\begin{figure}[!ht]
\begin{lstlisting}[style=prompt]
"Coupon Math Calculation Question":'''
You are a question-setting assistant. Your task is to set questions based on specific goals and rules to evaluate the instruction-following ability of the large language model.
First, here is the goal of the task:
<objective>{obj}</objective>

And, you need to follow the following {upper} rules when judging:
<rules>{rules}</rules>

The generated questions are as follows, avoiding too much repetition:
<questions>{question}</questions>

Now you need to set a coupon calculation question based on <rules>.
The requirements are as follows:
- For e-commerce, the generated content must include the background of the question, the coupons allowed to be used (mentioned in the rules), the variables to be calculated, etc.
- The copy must follow at least {lower} non-contradictory rules.
- Give reference answers and step-by-step methods to get the answer.

The questions need to be output to a list in json format. Each question contains 3 fields:
1. "question" field: fill in the copy content here, in string form, without any results and analysis.
2. "label" field: fill in the reference answer here, in the form of numbers or strings.
3. "analysis" field: for the specific method of judging the reference answer, just explain the content directly, and do not mention the rule_id based on the rule.
4. "picked_rules" field: a JSON list containing {lower} rules, retaining the "type", "rule_id", and "rule_text" of the selected rule.

Now please ask questions according to the requirements.
\end{lstlisting}
\caption{Prompt for math.}
\label{fig:context-4}
\end{figure}

\begin{figure}[!ht]
\begin{lstlisting}[style=prompt]
You are a question-setting assistant. Your task is to set questions based on specific goals and rules to evaluate the instruction-following ability of the large language model.

First, here is the goal of the task:
<objective>{obj}</objective>

And, you need to follow the following {upper} rules when judging:
<rules>{rules}</rules>

The generated questions are as follows, avoiding too much repetition:
<questions>{question}</questions>

Now you need to generate 2 open dialogues based on <rules>. In one open dialogue, the customer service strictly follows the given <rules>, and in the other open dialogue, the customer service does not fully comply with all the given <rules>.

The requirements are as follows:
- In the airline flight booking customer service scenario, the generated content should determine the user situation and goals based on the order information and the conversation content, and generate the corresponding user's reply speech requirements.
- Including both the user [user] and the customer service [assistant], the customer service reply includes at least 25 characters, including complex sentence structures.
- At least 2 rounds of dialogue between the user and the assistant, more rounds are possible.
- User questions are more colloquial

The questions need to be output to a list in json format, and each question contains 3 fields:

1. "question" field: fill in the dialogue, in string form, without any results and analysis.

2. "label" field: there are only 2 results, 0 represents the question that does not follow the customer service rules, and 1 represents the question that finally follows the customer service rules

3. "analysis" field: a brief reason for the judgment, just explain the content directly, and do not mention the rule_id based on the rule.

4. "picked_rules" field: a JSON list containing {lower} rules, retaining the "type", "rule_id", and "rule_text" of the selected rule.

Now please ask questions according to the requirements.
\end{lstlisting}
\caption{Prompt for agent chatting.}
\label{fig:context-6}
\end{figure}

\begin{figure}[!ht]
\begin{lstlisting}[style=prompt]
"Generate a summary of a general text according to specific rules":'''
You are a question-setting assistant, and your task is to set questions according to specific goals and rules to evaluate the instruction-following ability of a large language model.
First, here is the goal of the task:
<objective>{obj}</objective>

And, you need to follow the following {upper} rules when judging:
<rules>{rules}</rules>

The generated questions are as follows, avoiding too much repetition:
<questions>{question}</questions>

Now you need to give a summary question according to <rules>.

The requirements are as follows:
- The themes and content of the text need to be diverse and repetitions should be reduced.
- The text should be more than 250 characters and contain multiple short sentences and complex syntactic structures.
- The text must follow at least {lower} non-contradictory rules.
- Give a reference summary.

The questions need to be output to a list in json format. Each question contains 3 fields:
1. "question" field: fill in the text content here, in string format, excluding any result analysis.
2. "label" field: fill in the reference answer given here, in string format.
3. "analysis" field: for the analysis of the reasons for judging <objective>, just explain the content directly, and do not mention the rule_id based on the rule.
4. "picked_rules" field: a JSON list containing {lower} rules, retaining the "type", "rule_id", and "rule_text" of the selected rule.
Now please ask questions according to the requirements.
\end{lstlisting}
\caption{Prompt for summarization.}
\label{fig:context-3}
\end{figure}

\begin{figure}[!ht]
\begin{lstlisting}[style=prompt]
"Judge whether the content generated by the large model has hallucination problems and classify them":"""You are a question-setting assistant, responsible for setting questions based on specific goals. These questions are used to evaluate the instruction-following ability of the large language model.
First, the goal of the task is as follows:
<objective>{obj}</objective>

And, you need to follow the following {upper} rules when judging:
<rules>{rules}</rules>

The generated questions are as follows, avoiding too much repetition:
<questions>{question}</questions>

The requirements for setting questions are as follows:
1. A question with the judgment result of "no hallucination of the large model":
- The content of the question should be based on real life, and the logic, inference and grammatical expression should be correct.
- The content should be carefully constructed, and under normal logic and cognition, the content should not be wrong or unreasonable.

2. A question with the judgment result of "hallucination of the large model":
- The topics can be diverse, such as e-commerce, life, health, and knowledge. It is necessary to reflect the multiple types of errors in <rules>.
- The generated content must meet at least {lower} non-contradictory rules.
- Carefully construct errors. Errors should not be too obvious. They should not contain content that can be immediately judged as false or wrong, such as "magic", "mystery", "aliens", and "superpowers".
- Do not include analysis of errors, such as "some companies mistakenly believe" or "but in fact, this is a logical error" or "no factual basis" or "no data source" or "however, there is no basis".

3. Output multiple questions in a list, and output each question in json format. Each question contains 4 fields:
- "question" field: fill in the generated content in string format, excluding any result analysis. The question content must be more than 50 characters and contain multiple short sentences or complex syntactic structures.
- "label" field: There are only 2 results, 0 represents the question with the final judgment of "no hallucination", and 1 represents the question with the final judgment of "hallucination".
- "analysis" field: for the analysis of the reasons for judging <objective>, just explain the content directly, and do not mention the rule_id based on the rule.
- "picked_rules" field: a JSON list containing {lower} rules, retaining the "type", "rule_id", and "rule_text" of the selected rule.

Now, please ask questions according to the above requirements.
\end{lstlisting}
\caption{Prompt for hallucination detection.}
\label{fig:context-2}
\end{figure}

\subsection{Prompts for Select Relevant Rules}\label{app:grr}
The following are prompts for selecting the most relevant rules, as shown in Figure \ref{fig:ans-1}.

\FloatBarrier

\begin{figure}[!ht]
\begin{lstlisting}[style=prompt]
You are a big model assistant who is very good at semantic understanding. Your task is to select the most relevant {n} rules from the provided rule set according to the given target task.
### Target task
The target task is: {objective}
### Rule set
The following is the content of the rule set {rules}. Please extract the {n} rules that are most relevant to the target task. Each rule in the rule set includes "type" (rule type), "rule_id" (rule number), and "rule_text" (rule content). Please keep these fields and their contents intact.
### Requirements
1. According to the requirements of the target task, carefully analyze the semantics of each rule to ensure that the selected rule matches the target task.
2. The contents of the "type", "rule_id" and "rule_text" fields of the rule must not be modified. Only the most relevant rules can be selected and retained.
3. Make sure that the selected rules do not contradict or conflict with each other. For example, in the task of judging the same product, if rule 1 is "if the prescribed usage time is different, it is judged as a different product", and rule 2 is "if the prescribed usage time is different, it is judged as the same product", then rule 1 and rule 2 cannot be selected at the same time.
4. On the basis of 3, try to reflect diversity.
5. Output the results directly without any additional explanation or comment.
### Output format
The result is a JSON list containing {n} rules, in the following format:
[
{{
"type": "Rule type",
"rule_id": "Rule number",
"rule_text": "Rule content"
}},
...
]
Please output the results directly according to the above instructions without asking or requesting more information.
\end{lstlisting}
\caption{Prompt for selecting relevant rules.}
\label{fig:ans-1}
\end{figure}

\subsection{Prompts for Editing Rules}\label{app:edit_rules}
As shown in Figure \ref{fig:rules-1}, the prompt guides the modification of rule-based judgments by changing tag values. The updated rule is stored in \texttt{edit\_rule} with key attributes, and the new label is recorded in \texttt{edit\_label}, while maintaining the original JSON structure.

\begin{figure}[!ht]
\begin{lstlisting}[style=prompt]
According to the given JSON format question, complete the following tasks:
1. Prioritize a rule in rule-set or picked_rules, and adjust the content of the rule so that the label of the question changes (from 0 to 1 or from 1 to 0).
2. If there is no suitable rule in rule-set and picked_rules, you can add or modify the content of a rule to achieve the goal.
3. Store the adjusted rule in a new field `edit_rule`, and keep the following three contents:
- `type`: the type of the adjustment rule.
- `rule_id`: the unique identifier of the adjustment rule.
- `rule_text`: the text of the adjusted rule.
4. Store the adjusted `label` value in a new field `edit_label`.
**Enter JSON field description**:
- `question`: the product description and question context in the question.
- `label`: the original judgment result.
- `analysis`: the reason analysis for judging <objective>.
- `picked_rules`: currently selected rule list.
- `rule_set`: list of all rules, which can be adjusted or added.
- `objective`: judgment target.
**Input example:**
{{
"question": "{question}",
"label": {label},
"analysis": {analysis},
"picked_rules": {picked_rules},
"rule_set": {rule_set},
"objective": "{objective}"
}}
**Output requirements**:
Output the modified JSON content, including the original fields and the newly added fields `edit_rule` and `edit_label`.
\end{lstlisting}
\caption{Prompt for selecting relevant rules.}
\label{fig:rules-1}
\end{figure}

\subsection{Prompts for Generating Questions}\label{app:gq}
The following are prompts for generating questions and answer options, as shown in Figure \ref{fig:answer-1} \textendash{} \ref{fig:answer-3}.

\FloatBarrier

\begin{figure}[!ht]
\begin{lstlisting}[style=prompt]
You are an assistant who generates instruction compliance test questions. Your task goal is to generate 4 options of different quality based on the given instructions <guidelines> and the copy <context>, and then merge the instructions, copy, and options into a question to test instruction compliance ability, and output it in json format.
The following is the overall goal of the task:
<objective>{obj}</objectives>

The following are specific instructions:
<guidelines>{picked_rules}</guidelines>

The following is the given copy:
<context>{question}</context>

The generated options need to meet the following requirements:
1. The quality is clearly distinguished, and there is only one best option. Generate the best option based on <guidelines> and fill in the "OptimalOption" field of "Groundtruth".
2. Use A.B.C.D as the serial number, and separate different options with line breaks.
3. Each option needs to be able to directly answer <objective> and cannot be too brief.
4. The best option can refer to {label}, the reason is {analysis}

The final output json format should contain the following fields:

"Instruction": {obj},

"Guidelines": {picked_rules},

"Context": {question},

"MultipleOptions": 4 generated options, string format,

"Groundtruth":

"OptimalOption": "Best option, only output option number A/B/C/D",

"AnswerAnalysis": Analyze the reasons for choosing the best option, briefly analyze the reasons why other options do not meet <guidelines>, no need to mention rule_id and type

Please generate questions as required.
\end{lstlisting}
\caption{Prompt for generating answer options.}
\label{fig:answer-1}
\end{figure}

\begin{figure}[!ht]
\begin{lstlisting}[style=prompt]
You are an assistant who generates instruction compliance test questions. Your task goal is to combine the given instructions <guidelines> and copy <context>, the best answer and the reason for the best answer into a question to test the ability to follow instructions, and output it in json format.
The following is the overall objective of the task:
<objective>{obj}</objectives>

The following is the specific instruction:
<guidelines>{picked_rules}</guidelines>

The following is the given copy:
<context>{question}</context>

Best answer:
<groundtruth>{label}</groundtruth>

Reason for generating the best answer:
<groundtruth_analysis>{analysis}</groundtruth_analysis>

The final output json format should contain the following fields:
"Instruction": {obj},
"Guidelines": {picked_rules},
"Context": {question},
"Groundtruth":
"ReferenceAnswer": {label},
"ReferenceAnalysis": {anaylysis}
Please generate the question as required.
\end{lstlisting}
\caption{Prompt for combining the instructions and context.}
\label{fig:answer-2}
\end{figure}

\begin{figure}[!ht]
\begin{lstlisting}[style=prompt]
Your task is to select the best option and give a brief analysis based on the instructions, copywriting and candidate options provided.
First, please understand carefully:
Instruction is a general task goal, and its content is:
<Instruction>{Instruction}</Instruction>
Guideline is a user-defined rule, and its content is:
<Guidelines>{Guidelines}</Guidelines>

Copywriting content:
<Context>{Context}</Context>

Candidate options:
<Options>{MultipleOptions}</Options>

When evaluating options, you need to pay attention to:
1. Carefully compare the degree of compliance, accuracy and completeness of each option with the instructions <Instruction> and <Guidelines>.

Output in json format, including two fields, OptimalOption and AnswerAnalysis. Among them, OptimalOption only outputs the best option number, and AnswerAnalysis gives a brief analysis. Do not point out rule_id and type, and briefly compare the difference between the best option and other options.
\end{lstlisting}
\caption{Prompt for selecting the best option.}
\label{fig:answer-3}
\end{figure}

\subsection{Prompt Templates for Guiding the Model to Answer Questions}\label{app:aq}
The following are prompts for generating questions and answer options, as shown in Figure \ref{fig:aq-1} \textendash{} \ref{fig:aq-4}.

\FloatBarrier
\begin{figure}[!ht]
\begin{lstlisting}[style=prompt]
Your task is to output the judgment results and specific analysis based on the provided instructions and copywriting.
First, please understand carefully:
Instruction is a general task goal, the content is:
<Instruction>{Instruction}</Instruction>
Guideline is a user-defined rule, the content is:
<Guidelines>{Guidelines}</Guidelines>
Copywriting content:
<Context>{Context}</Context>
Output in json format. Includes two fields, CandidateAnswer and CandidateAnalysis. Among them, CandidateAnswer only outputs 0 or 1, 0 represents negative, 1 represents positive; CandidateAnalysis gives a brief analysis, do not specify rule_id and type.
\end{lstlisting}
\caption{Prompt template for audit algorithm and price matching questions.}
\label{fig:aq-1}
\end{figure}

\begin{figure}[!ht]
\begin{lstlisting}[style=prompt]
Your task is to select the best option and give a brief analysis based on the instructions, copywriting and candidate options provided.
First, please understand carefully:
Instruction is a general task goal, and its content is:
<Instruction>{Instruction}</Instruction>
Guideline is a user-defined rule, and its content is:
<Guidelines>{Guidelines}</Guidelines>
Copywriting content:
<Context>{Context}</Context>
Candidate options:
<Options>{MultipleOptions}</Options>
When evaluating options, you need to pay attention to:
1. Carefully compare the degree of compliance, accuracy and completeness of each option with the instructions <Instruction> and <Guidelines>.
Output in json format. Includes two fields, OptimalOption and AnswerAnalysis. Among them, OptimalOption only outputs the best option number, and AnswerAnalysis gives a brief analysis. Do not point out rule_id and type, and briefly compare the difference between the best option and other options.
\end{lstlisting}
\caption{Prompt template for multiple choice questions.}
\label{fig:aq-2}
\end{figure}

\begin{figure}[!ht]
\begin{lstlisting}[style=prompt]
Your task is to output the judgment results and specific analysis according to the provided instructions and copywriting.
First, please understand carefully:
Instruction is a general task goal, the content is:
<Instruction>{Instruction}</Instruction>
Guideline is a user-defined rule, the content is:
<Guidelines>{Guidelines}</Guidelines>
Copywriting content:
<Context>{Context}</Context>
Output in json format. Includes two fields, CandidateAnswer and CandidateAnalysis. Among them, CandidateAnswer only outputs one of the following three judgment results, "2 (strong correlation)", "1 (weak correlation)", "0 (irrelevant)"; CandidateAnalysis gives the necessary analysis process, string format, do not specify rule_id and type.
\end{lstlisting}
\caption{Prompt template for text relevance questions.}
\label{fig:aq-4}
\end{figure}

\begin{figure}[!ht]
\begin{lstlisting}[style=prompt]
Your task is to output the judgment results and specific analysis based on the provided instructions and copywriting.
First, please understand carefully:
Instruction is a general task goal, the content is:
<Instruction>{Instruction}</Instruction>
Guideline is a user-defined rule, the content is:
<Guidelines>{Guidelines}</Guidelines>
Copywriting content:
<Context>{Context}</Context>
Output in json format. Includes two fields, CandidateAnswer and CandidateAnalysis. Among them, CandidateAnswer only outputs the calculation result, in string format, without the unit "yuan"; CandidateAnalysis gives the necessary calculation process, in string format, do not specify rule_id and type.
\end{lstlisting}
\caption{Prompt template for math questions.}
\label{fig:aq-3}
\end{figure}

\FloatBarrier
\subsection{Prompt for Extracting JSON Answer}\label{Extracting}
The answers generated by 7b models such as Qwen2.5-7B often do not follow the JSON format, so they cannot be automatically compiled into the JSON format for evaluation. To solve this problem, we used gpt-4o to analyze the results of the model, extract the answers, and organize them into JSON format for return. The prompts for extracting answers from different domains are as shown in Figure \ref{fig:extract-1} \textendash{} \ref{fig:extract-3}.

\begin{figure*}[!ht]
\begin{lstlisting}[style=prompt]
Please extract the answer ("CandidateAnswer") and analysis ("CandidateAnalysis") from the following text respectively. The returned content must have and only have two fields, "CandidateAnswer" and "CandidateAnalysis". Among them, CandidateAnswer only outputs 0 or 1 of int type, 0 represents negative, and 1 represents positive; the output json format is as follows:
{
"CandidateAnswer": "",
"CandidateAnalysis": ""
}
If a field has no content, please return an empty string. Note: Only json content is returned.
\end{lstlisting}
\caption{Prompt for extracting the answer of price matching and audit algorithm.}
\label{fig:extract-1}
\end{figure*}

\begin{figure*}[!ht]
\begin{lstlisting}[style=prompt]
Please extract the answer ("CandidateAnswer") and analysis ("CandidateAnalysis") in the following text respectively. The returned content must have and only have two fields, "CandidateAnswer" and "CandidateAnalysis". CandidateAnswer only outputs one of the following three judgment results, "2 (strong correlation)", "1 (weak correlation)", "0 (irrelevant)"; the output json format is as follows:
{
"CandidateAnswer": "",
"CandidateAnalysis": ""
}
If a field has no content, please return an empty string. Note: Only json content is returned.
\end{lstlisting}
\caption{Prompt for extracting the answer of text relevance calculation.}
\label{fig:extract-2}
\end{figure*}

\begin{figure*}[!ht]
\begin{lstlisting}[style=prompt]
Please extract the answer ("OptimalOption") and analysis ("AnswerAnalysis") from the following text, and return them as the values of the two fields "OptimalOption" and "AnswerAnalysis" in json format. OptimalOption only outputs the option number. Note: only json content is returned.
The output json format is as follows:
{
"OptimalOption": "",
"AnswerAnalysis": ""
}
If a field has no content, please return an empty string.
\end{lstlisting}
\caption{Prompt for extracting the answer of text relevance, hallucination detection, math, and agent chatting
.}
\label{fig:extract-3}
\end{figure*}

\FloatBarrier
\subsection{Prompts with Basic Instruction and Context but without Guidelines}\label{app:without_guide}
In this section, we provide the prompt used in our ablation study experiment with GPT-4o*. This version incorporates only the basic instruction and context without additional guidelines. The specific prompts used in our experiment are shown in Figure \ref{fig:without_guide-1} \textendash{} \ref{fig:without_guide-4}.

\begin{figure*}[!ht]
\begin{lstlisting}[style=prompt]
Your task is to output the judgment results and specific analysis based on the provided instructions and text.
First, please understand carefully:
Instruction is a general task target, and the content is:
<Instruction>{Instruction}</Instruction>
Text content:
<Context>{Context}</Context>
Output in json format. Includes two fields, CandidateAnswer and CandidateAnalysis. Among them, CandidateAnswer only outputs the calculation result, in string format, without the unit "yuan"; CandidateAnalysis gives the necessary calculation process, in string format, and does not specify rule_id and type.
\end{lstlisting}
\caption{Prompt without guidelines for audit algorithm and price matching questions.}
\label{fig:without_guide-1}
\end{figure*}

\begin{figure*}[!ht]
\begin{lstlisting}[style=prompt]
Your task is to output the judgment results and specific analysis based on the provided instructions and copy.
First, please understand carefully:
Instruction is a general task goal, and the content is:
<Instruction>{Instruction}</Instruction>
Copy content:
<Context>{Context}</Context>
Output in json format. Includes two fields, CandidateAnswer and CandidateAnalysis. Among them, CandidateAnswer only outputs 0 or 1, 0 represents negative, and 1 represents positive; CandidateAnalysis gives a brief analysis, and do not specify rule_id and type.
\end{lstlisting}
\caption{Prompt without guidelines for multiple choice questions.}
\label{fig:without_guide-2}
\end{figure*}

\begin{figure*}[!ht]
\begin{lstlisting}[style=prompt]
Your task is to select the best option and give a brief analysis based on the instructions, copywriting and candidate options provided.
First, please understand carefully:
Instruction is a general task goal, and its content is:
<Instruction>{Instruction}</Instruction>
Copywriting content:
<Context>{Context}</Context>
Candidate options:
<Options>{MultipleOptions}</Options>
When evaluating options, you need to pay attention to:
1. Carefully compare the degree of compliance, accuracy and completeness of each option with the instruction <Instruction>.
Output in json format. Includes two fields, OptimalOption and AnswerAnalysis. Among them, OptimalOption only outputs the best option number, and AnswerAnalysis gives a brief analysis. Do not point out rule_id and type, and briefly compare the difference between the best option and other options.
\end{lstlisting}
\caption{Prompt without guidelines for text relevance questions.}
\label{fig:without_guide-3}
\end{figure*}

\begin{figure*}[!ht]
\begin{lstlisting}[style=prompt]
Your task is to output the judgment results and specific analysis according to the provided instructions and copy.
First, please understand carefully:
Instruction is a general task goal, the content is:
<Instruction>{Instruction}</Instruction>
Copy content:
<Context>{Context}</Context>
Output json format. Includes two fields, CandidateAnswer and CandidateAnalysis. Among them, CandidateAnswer only outputs one of the following three judgment results, "2 (strong correlation)", "1 (weak correlation)", "0 (irrelevant)"; CandidateAnalysis gives the necessary analysis process, string format, do not specify rule_id and type.
\end{lstlisting}
\caption{Prompt without guidelines for math questions.}
\label{fig:without_guide-4}
\end{figure*}

\FloatBarrier
\subsection{Prompt for Direct Output without COT Reasoning}
\label{app:re_evaluation_prompt}

In this section, we provide the prompts used in our re-evaluation of summarization and math tasks. This experiment aims to analyze the impact of Chain of Thought (CoT) by comparing model performance with and without CoT prompting.

Unlike the previous experiments, we directly output results without using CoT to examine the raw model capabilities. The specific prompts for math and summarization tasks are shown in Figure \ref{fig:re_evaluation_prompt-1} \textendash{} \ref{fig:re_evaluation_prompt-3}.



\begin{figure*}[!ht]
\begin{lstlisting}[style=prompt]
Your task is to output the judgment results and specific analysis based on the provided instructions and text.
First, please understand carefully:
Instruction is a general task goal, the content is:
<Instruction>{Instruction}</Instruction>
Guideline is a user-defined rule, the content is:
<Guidelines>{Guidelines}</Guidelines>
Text content:
<Context>{Context}</Context>
Output in json format. Includes 1 field, CandidateAnswer. CandidateAnswer must only directly output the calculation result, in string format, without the unit "yuan", and do not output the process of obtaining the question.
\end{lstlisting}
\caption{Prompt for direct output without COT in math task.}
\label{fig:re_evaluation_prompt-1}
\end{figure*}


\begin{figure*}[!ht]
\begin{lstlisting}[style=prompt]
Your task is to select the best option and give a brief analysis based on the provided instructions, copywriting, and candidate options.
First, please understand carefully:
Instruction is a general task goal, and its content is:
<Instruction>{Instruction}</Instruction>
Guideline is a user-defined rule, and its content is:
<Guidelines>{Guidelines}</Guidelines>
Copywriting content:
<Context>{Context}</Context>
Candidate options:
<Options>{MultipleOptions}</Options>
When evaluating options, please note:
1. Carefully compare the degree of compliance, accuracy, and completeness of each option with the instruction <Instruction>.
Output in json format. Includes 1 field, OptimalOptios. Among them, OptimalOption only outputs the best option sequence number, a string, and does not output any other analysis.
\end{lstlisting}
\caption{Prompt for direct output without COT in summarization task.}
\label{fig:re_evaluation_prompt-3}
\end{figure*}


\FloatBarrier
\section{Raw Results}
\label{app:raw_results}

During experiments, we observed that many baseline models, particularly those with smaller parameter sizes, failed to strictly adhere to the required output format. The lack of consistency significantly impacted the experimental results seen in Table~\ref{tab:not_parsed_main_results}.

\begin{table*}[htb]
  \centering
  \renewcommand\tabcolsep{4pt} 
  \scalebox{0.65}{
  \begin{tabular}{l|c|ccccccc}
  \toprule
  \multirow{2}{*}{Models} & \multirow{1}{*}{\textit{All}} & \multicolumn{7}{c}{\textit{Task Categories}} \\
    \cmidrule{2-2}
    \cmidrule(lr){3-9}
    & Accuracy & audit algorithm & price matching &  text relevance & math & agent chatting & summarization &  hallucination detection \\\midrule
    o1  & 79.17  & 73.48  & 76.24 & 79.69  & \underline{48.08}  & 92.78 & 81.03 & 92.37 \\
    GPT-4o & \underline {86.48} & \underline{96.52}  & \underline{84.84}  & 81.25  & 13.46  & \textbf{100} & \underline{82.76} & 94.92  \\
     
    GPT-4o* & 80.90 & 94.78  & 74.66  & 80.21  & 7.69  & 95.56 & 68.97 & 94.07  \\\midrule
    
   Deepseek-R1  & \textbf{87.26}  & 93.04  & 80.32 & 84.90 & \textbf{65.38} & \underline{98.89}  & \textbf{89.66 } & \textbf{96.61} \\
    Deepseek-V3 & 83.96 & \textbf{97.39}  & \textbf{91.18}  & 53.65  & 5.77 & \underline{98.89} & 77.59 & 94.92  \\\midrule

    Gemini-2.5-pro-exp & 75.63 & 90.00 & 75.79 & \underline{85.94} & 44.23 & 80.00 & 70.69 & 39.83 \\ \midrule

    Mistral-7B-Instruct & 56.21 & 80.43 & 66.06 & 4.69 & 1.92 & 58.33 & 31.03 & 88.98 \\ \midrule

    Yi-1.5-6B & 0.08 & 0.44 & 0.00 & 0.00 & 0.00 & 0.00 & 0.00 & 0.00 \\
    Yi-1.5-34B & 0.00 & 0.00 & 0.00 & 0.00 & 0.00 & 0.00 & 0.00 & 0.00 \\ \midrule

    Gemma-3-4b-it & 58.02 & 55.65 & 55.88 & 54.69 & 0.00 & 76.67 & 72.41 & 66.10 \\ \midrule
        
    QwQ-32B-Preview & 38.84 & 29.57 & 14.93 & 30.73 & 3.85 & 86.67 & 62.07 & 90.68 \\

    R1-Distill-Qwen-7B  & 0.00 & 0.00 & 0.00  & 0.00  & 0.00  & 0.00  & 0.00  & 0.00 \\
    
    R1-Distill-Qwen-32B  & 0.00 & 0.00 & 0.00  & 0.00  & 0.00  & 0.00  & 0.00  & 0.00 \\

    Qwen2.5-7B  & 0.00  & 0.00  & 0.00  & 0.00 & 0.00 & 0.00  & 0.00  & 0.00 \\
    Qwen2.5-7B-Instruct  & 25.08  & 0.00 & 0.00 & 0.00  & 0.00  & 95.56  & 67.24  & 91.53 \\

    Qwen2.5-32B  & 9.40  & 0.00  & 0.00 & 0.00 & 0.00  & 0.00  & 18.97 & 0.85 \\\midrule
    
    Vicuna-7B  & 6.37  & 29.57  & 2.04 & 2.08 & 0.00 & 0.00 & 0.00 & 0.00 \\\midrule
    
    Llama3-8B-Instruct  & 0.00  & 0.00  & 0.00  & 0.00  & 0.00 & 0.00 & 0.00  & 0.00 \\
    Llama-3.3-70B-Instruct & 86.08 & \textbf{97.39} & 82.58 & \textbf{86.98} & 7.69 & 96.67 & \underline{82.76} & \underline{95.76} \\

    \bottomrule
  \end{tabular}}
  \caption{Main results (\%) of different models across 7 task categories. Raw answers without processing. Accuracy is computed based on the number of correctly predicted labels. Segment 1: GPT series; Segment 2: Deepseek series; Segment 3: Gemini series; Segment 4: Mistral series; Segment 5: Yi series; Segment 6: QwQ series; Segment 7: Gemma series; Segment 8: Qwen series; Segment 9: Vicuna series; Segment 10: Llama series. The best-performing results are shown in \textbf{bold} face, and the second-best are \underline{underlined}. GPT-4o* denotes the ablation study using GPT-4o with basic \textit{Instruction} and \textit{Context}, but without \textit{Guidelines}.}
  \label{tab:not_parsed_main_results}
\end{table*}

\FloatBarrier
\section{Error Analysis}
\label{app:error_analysis}

An example of logical error is illustrated in Figure~\ref{fig:part1_logic} and Figure~\ref{fig:part2_logic}, where the error stems from the model's failure to comprehend the guideline rule that 25 yuan fixed amount coupon can be combined with discount coupons but cannot be combined with full reduction coupons.
Besides, the example of commonsense error is illustrated in Figure~\ref{fig:commonerror}, where the error stems from the model's misunderstanding about the rule that discount coupon can be applied to home goods.

\begin{figure*}[!ht]
\begin{lstlisting}[style=prompt, escapeinside={(*@}{@*)}]
{
    "Groundtruth": {
      "ReferenceAnswer": "1303",
      "ReferenceAnalysis": "For the 600-yuan non-food, non-beverage home goods (non-sports equipment), since today is Saturday, both the 10% off coupon and the 30% off coupon can be used, resulting in a discounted price of 600*0.9*0.7=378 yuan. For the 950-yuan non-food, non-beverage, non-beauty, and non-home goods, the Spend 900 yuan and Save 220 yuan coupon can be used, resulting in a discounted price of 950 - 220 = 730 yuan. However, the two discounts cannot be used simultaneously, and the 25-yuan coupon can be used together with the discount coupons. Choosing the discount coupons, the final payment is 378 + 950 - 25 = 1303 yuan. Choosing the full reduction would require a payment of 600 + 730 = 1330 yuan. Therefore, choosing the lesser amount, 1303 yuan, is better."
    },
    "LLMAnswer": {
      "CandidateAnswer": "1305",
      "CandidateAnalysis": "Xiao Zhang's order includes 950 yuan worth of non-home goods and 600 yuan worth of home goods. First, use the full reduction coupon of Spend 900, Save 220 (assuming valid usage time), so the net price of the 950-yuan order is reduced to 730 yuan. The home goods section cannot stack a discount and remains at 600 yuan. (*@\textcolor{blue}{After applying the fixed amount }@*)
      (*@\textcolor{blue}{coupon of 25 yuan, the total amount to pay is 730 + 600 - 25 = 1305 yuan.}@*)"
    }
}
\end{lstlisting}
\caption{Part 2 of the logical error of Deepseek-R1 response in a math problem. The error step has been highlighted in blue: the 25 yuan fixed amount coupon cannot be used with the full reduction coupon.}
\label{fig:part2_logic}
\end{figure*}

\begin{figure*}[!ht]
\begin{lstlisting}[style=prompt]
{
    "Instruction": "Coupon Math Problem",
    "Guidelines": [
      {
        "type": "Discount Coupon",
        "rule_text": "Discount Coupon: 30% off coupon, applicable to all home goods except sports equipment, limited to one use per order. Can be combined with full reduction coupons and fixed amount coupons."
      },
      {
        "type": "Fixed Amount Coupon",
        "rule_text": "Fixed Amount Coupon: 25 yuan coupon, no minimum spending requirement, applicable to all products except food, beverages, and beauty products. Limited to one use per order. Can be combined with discount coupons but cannot be combined with full reduction coupons."
      },
      {
        "type": "Full Reduction Coupon",
        "rule_text": "Full Reduction Coupon: Spend 700 yuan and save 180 yuan. Can be combined with fixed amount coupons but cannot be combined with discount coupons. Limited to one use per order. Applicable to all products except food, beverages, electronics, and sports equipment."
      },
      {
        "type": "Combined Coupon Usage Restriction",
        "rule_text": "In the same order, if both full reduction coupons and fixed amount coupons are used, the full reduction amount must be calculated first, followed by the deduction amount from the fixed amount coupon."
      },
      {
        "type": "Fixed Amount Coupon",
        "rule_text": "Fixed Amount Coupon: New 12-yuan coupon, no minimum spending requirement, applicable to toy products on the platform. No time restrictions. Limited to one use per order. Can be combined with full reduction coupons and discount coupons."
      },
      {
        "type": "Discount Coupon",
        "rule_text": "Discount Coupon: New 35% off coupon, applicable to sports equipment products on the platform. Valid during the middle of each month. Limited to one use per order. Cannot be combined with full reduction coupons but can be combined with fixed amount coupons."
      },
      {
        "type": "Combined Coupon Usage Restriction",
        "rule_text": "In the same order, if multiple types of non-stackable coupons are used, only one of them can be selected."
      },
      {
        "type": "Discount Coupon",
        "rule_text": "Discount Coupon: New 10% off coupon, applicable to all home goods except food and beverages. Valid from Friday to Sunday each week. Limited to one use per order. Can be combined with full reduction coupons and fixed amount coupons."
      },
      {
        "type": "Full Reduction Coupon",
        "rule_text": "Full Reduction Coupon: New Spend 900 yuan and save 220 yuan. Can be combined with fixed amount coupons but cannot be combined with discount coupons. Valid during the end of each month. Applicable to all products except beauty products and home goods."
      }
    ],
    "Context": "When shopping on an e-commerce platform, Xiao Zhang plans to purchase a batch of items, including 950 yuan worth of non-food, non-beauty, and non-home goods and 600 yuan worth of non-food home goods (non-sports equipment). The platform offers various types of coupons are listed above. Assuming today is Saturday, Xiao Zhang wants to maximize his discounts. How much does he need to pay in the end?"
  }
}
\end{lstlisting}
\caption{Part 1 of the logical error of Deepseek-R1 response in a math problem.}
\label{fig:part1_logic}
\end{figure*}

\begin{figure*}[!ht]
\begin{lstlisting}[style=prompt, escapeinside={(*@}{@*)}]
{
    "Instruction": "Coupon math problem",
    "Guidelines": [
      {
        "type": "Discount Coupon",
        "rule_text": "Discount Coupon: 30% off coupon, applicable to all home goods except sports equipment. Can only be used once per order and can be stacked with fixed amount coupons or tiered discount coupons."
      },
      {
        "type": "Fixed Amount Coupon",
        "rule_text": "Fixed Amount Coupon: 25 coupon with no spending threshold, applicable to all items on the platform except food and beverages or beauty products. Can only be used once per order, can stack with discount coupons but cannot stack with tiered discount coupons."
      },
      {
        "type": "Tiered Discount Coupon",
        "rule_text": "Tiered Discount Coupon: Spend 700 and get 180 off. Can be stacked with fixed amount coupons but cannot stack with discount coupons. Can only be used once per order, applicable to all items on the platform except food and beverages, electronics, and sports equipment."
      }
    ],
    "Context": "On an e-commerce platform, Xiao Wang wants to buy a home decor item (not sports equipment) priced at 750 and a general item (not food, beverages, beauty products, electronics, or sports equipment) priced at 50. The platform offers various coupons as above. How should Xiao Wang use the coupons to maximize savings, and what is the final amount to be paid?",
    "Groundtruth": {
      "ReferenceAnswer": "550",
      "ReferenceAnalysis": "First, analyze the usage scenarios for the three types of coupons. If the tiered discount coupon is used, the total price is 750 + 50 = 800. After a 180 discount for spending 700, the payment required is 800 - 180 = 620. If the discount coupon is used, the home decor item priced at 750 gets a 30% discount, making it 750*0.7 = 525. Adding the 50 general item, the total is 525 + 50 = 575. Then, using the 25 fixed amount coupon, the final payment required is 575 - 25 = 550. Comparing the options, using the discount coupon and fixed amount coupon is more cost-effective, with a final payment of 550."
    },
    "LLMAnswer": {
      "CandidateAnswer": "535",
      "CandidateAnalysis": "The total price of Xiao Wang's purchases is 750 (home decor) + 50 (general item) = 800. According to the coupon rules, the 30% discount coupon and the 25 fixed amount coupon can be stacked. (*@\textcolor{blue}{First, apply the 30\% discount coupon to the total price of 800: }@*)
      (*@\textcolor{blue}{800 * 0.7 = 560 yuan.}@*) Then, apply the 25 fixed amount coupon: 560 - 25 = 535. If the tiered discount coupon is used alone (800 - 180 = 620) or if other coupons are used separately, the cost would be higher than yuan535. Therefore, the optimal choice is to use the 30% discount coupon and 25 fixed amount coupon together, resulting in a final payment of 535 yuan."
    }
  }
\end{lstlisting}
\caption{Error of the commonsense error of Deepseek-R1 response in a math problem. The error step has been highlighted in blue: the 30\% discount coupon can only be applied to home goods instead of all goods.}
\label{fig:commonerror}
\end{figure*}

\FloatBarrier

\section{Examples of Over-thinking and Under-thinking}
\label{app:overandunder}
 Examples of over-thinking and under-thinking phenomena are respectively illustrated in Figure~\ref{fig:over-think} and Figure~\ref{fig:underthinking}.

\begin{figure}[!ht]
\begin{lstlisting}[style=prompt]
Okay, let's tackle this problem. So, the task is to determine if Product A and Product B are the same product ... Wait, no, the rule says if one mentions and the other doesn't, then they are different. But here, both mention it. So maybe rule 84 doesn't apply here... Wait, but the rule only states that if one mentions and the other doesn't, then they're different... Wait, but the rule says if one mentions and the other doesn't... Wait, but if the key features are present in both, but the parameters differ ... So perhaps differences in parameters (like 4K vs 8K) aren't covered by the given rules... However, wait, the rule_23 says if one mentions key features and the other doesn't... Wait, but ... Therefore, the answer is 1.
\end{lstlisting}
\caption{An illustration of over-thinking of price matching tasks.}
\label{fig:over-think}
\vspace{-5mm}
\end{figure}

\begin{figure}[!ht]
\begin{lstlisting}[style=prompt]
"Instruction": "Coupon Math Calculation Problem",
"Guidelines": [
  {
    "type": "Restrictions on combined coupon usage",
            "rule_id": "rule_34",
            "rule_text": "If one coupon states it cannot be stacked with a certain type of coupon while another states it can, the 'cannot be stacked' rule takes precedence"
  },
  ...
],
"Context": "On the weekend, Xiao Wang bought a computer priced at 1200 CNY (not belonging to mother and baby supplies, toys, or food and beverages) on an e-commerce platform, and also bought sporting goods priced at 300 CNY...  How should Xiao Wang use the coupons to achieve the lowest price, and how much does he have to pay in total?",
"LLMAnswer": {
  "CandidateAnswer": "1250",
  "ThinkingProcess":"Okay, let's try to figure out how Xiao Wang can use the coupons to get the lowest price. First, I need to break down his purchases and the available coupons...Let's compare: Option A: Use 250 off threshold coupon: 1500 -250 = 1250. Option B: Use 40 off on sporting goods: 300-40=260. Total 1200+260=1460. But can he combine with another coupon? The 40 off can't be combined with others, so no. So total 1460. Option C: Is there a way to split the order? Like apply the 250 off to the computer (1200), which is over 1000...Thus, the minimum payment is 1250."
}
\end{lstlisting}
    \caption{An illustration of under-thinking of math tasks.}
    \label{fig:underthinking}
\end{figure}

\end{document}